\documentclass[12pt]{article}
\usepackage{graphicx}

\usepackage{amsmath}
\usepackage{times}
\usepackage[dvipsnames]{xcolor}
\usepackage{hyperref}
\hypersetup{
    colorlinks=true,
    urlcolor=MidnightBlue,
    citecolor=PineGreen,
    linkcolor=RedOrange,
}

\usepackage{epigraph}
\usepackage{titletoc,tocloft}
\title{On the Measure of Intelligence} 

\author{Fran\c{c}ois Chollet
    \thanks{I thank Jos{\'e} Hern{\'a}ndez-Orallo, Julian Togelius, Christian Szegedy, and Martin Wicke for their valuable comments on the draft of this document.}\\
    \normalsize{Google, Inc.}\\
    \normalsize{\textit{fchollet@google.com}}
}

\date{November 5, 2019}

\begin{document}

\maketitle

\abstract{To make deliberate progress towards more intelligent and more human-like artificial systems, we need to be following an appropriate feedback signal: we need to be able to define and evaluate intelligence in a way that enables comparisons between two systems, as well as comparisons with humans. Over the past hundred years, there has been an abundance of attempts to define and measure intelligence, across both the fields of psychology and AI. We summarize and critically assess these definitions and evaluation approaches, while making apparent the two historical conceptions of intelligence that have implicitly guided them. We note that in practice, the contemporary AI community still gravitates towards benchmarking intelligence by comparing the \textit{skill} exhibited by AIs and humans at specific tasks, such as board games and video games. We argue that solely measuring skill at any given task falls short of measuring intelligence, because skill is heavily modulated by prior knowledge and experience: unlimited priors or unlimited training data allow experimenters to ``buy'' arbitrary levels of skills for a system, in a way that masks the system's own generalization power. We then articulate a new formal definition of intelligence based on Algorithmic Information Theory, describing intelligence as \textit{skill-acquisition efficiency} and highlighting the concepts of \textit{scope}, \textit{generalization difficulty}, \textit{priors}, and \textit{experience}, as critical pieces to be accounted for in characterizing intelligent systems. Using this definition, we propose a set of guidelines for what a general AI benchmark should look like. Finally, we present a new benchmark closely following these guidelines, the Abstraction and Reasoning Corpus (ARC), built upon an explicit set of priors designed to be as close as possible to innate human priors. We argue that ARC can be used to measure a human-like form of general fluid intelligence and that it enables fair general intelligence comparisons between AI systems and humans.
}

{\hypersetup{hidelinks}
    \tableofcontents
}

\section{Context and history}

\subsection{Need for an actionable definition and measure of intelligence}
\label{needForAnActionableDefinition}

The promise of the field of AI, spelled out explicitly at its inception in the 1950s and repeated countless times since, is to develop machines that possess intelligence comparable to that of humans. But AI has since been falling short of its ideal: although we are able to engineer systems that perform extremely well on specific tasks, they have still stark limitations, being brittle, data-hungry, unable to make sense of situations that deviate slightly from their training data or the assumptions of their creators, and unable to repurpose themselves to deal with novel tasks without significant involvement from human researchers.

If the only successes of AI have been in developing narrow, task-specific systems, it is perhaps because only within a very narrow and grounded context have we been able to \textit{define our goal} sufficiently precisely, and to \textit{measure progress} in an actionable way. Goal definitions and evaluation benchmarks are among the most potent drivers of scientific progress. To make progress towards the promise of our field, we need precise, quantitative definitions and measures of intelligence -- in particular human-like general intelligence. These would not be merely definitions and measures meant to describe or characterize intelligence, but precise, explanatory definitions meant to serve as a North Star, an objective function showing the way towards a clear target, capable of acting as a reliable measure of our progress and as a way to identify and highlight worthwhile new approaches that may not be immediately applicable, and would otherwise be discounted.

For instance, common-sense dictionary definitions of intelligence may be useful to make sure we are talking about the same concepts, but they are not useful for our purpose, as they are not actionable, explanatory, or measurable. Similarly, the Turing Test \cite{Turing1950} and its many variants (e.g. Total Turing Test and Loebner Prize \cite{powers-1998-total}) are not useful as a driver of progress (and have in fact served as a red herring \footnote{Turing's imitation game was largely meant as an argumentative device in a philosophical discussion, not as a literal test of intelligence. Mistaking it for a test representative of the goal of the field of AI has been an ongoing problem.}), since such tests completely opt out of objectively defining and measuring intelligence, and instead outsource the task to unreliable human judges who themselves do not have clear definitions or evaluation protocols.

It is a testimony to the immaturity of our field that the question of what we mean when we talk about intelligence still doesn't have a satisfying answer. What's worse, very little attention has been devoted to rigorously defining it or benchmarking our progress towards it. Legg and Hutter noted in a 2007 survey of intelligence definitions and evaluation methods \cite{LeggHutter2007}: \textit{``to the best of our knowledge, no general survey of tests and definitions has been published''}. A decade later, in 2017, Hern{\'a}ndez-Orallo released an extensive survey of evaluation methods \cite{HernandezOrallo2016} as well as a comprehensive book on AI evaluation \cite{MeasureOfAllMinds2017}. Results and recommendations from both of these efforts have since been largely ignored by the community.

We believe this lack of attention is a mistake, as the absence of widely-accepted explicit definitions has been substituted with implicit definitions and biases that stretch back decades. Though invisible, these biases are still structuring many research efforts today, as illustrated by our field's ongoing fascination with outperforming humans at board games or video games (a trend we discuss in \ref{currentTrendsInAIEvaluation} and \ref{criticalAssessment}). The goal of this document is to point out the implicit assumptions our field has been working from, correct some of its most salient biases, and provide an actionable formal definition and measurement benchmark for human-like general intelligence, leveraging modern insight from developmental cognitive psychology.

\subsection{Defining intelligence: two divergent visions}
\label{definingIntelligence}

\epigraph{Looked at in one way, everyone knows what intelligence is; looked at in another way, no one does.}{Robert J. Sternberg, 2000}

Many formal and informal definitions of intelligence have been proposed over the past few decades, although there is no existing scientific consensus around any single definition. Sternberg \& Detterman noted in 1986 \cite{SternbergDetterman1986} that when two dozen prominent psychologists were asked to define intelligence, they all gave somewhat divergent answers. In the context of AI research, Legg and Hutter \cite{LeggHutter2007} summarized in 2007 no fewer than 70 definitions from the literature into a single statement: \textit{``Intelligence measures an agent's ability to achieve goals in a wide range of environments.''}

This summary points to two characterizations, which are nearly universally -- but often separately -- found in definitions of intelligence: one with an emphasis on task-specific skill (\textit{``achieving goals''}), and one focused on generality and adaptation (\textit{``in a wide range of environments''}). In this view, an intelligent agent would achieve high skill across many different tasks (for instance, achieving high scores across many different video games). Implicitly here, the tasks may not necessarily be known in advance: to truly achieve generality, the agent would have to be able to learn to handle new tasks (skill acquisition).

These two characterizations map to Catell's 1971 theory of fluid and crystallized intelligence (Gf-Gc) \cite{Cattell1971}, which has become one of the pillars of the dominant theory of human cognitive abilities, the Cattell-Horn-Caroll theory (CHC) \cite{CHC}. They also relate closely to two opposing views of the nature of the human mind that have been deeply influential in cognitive science since the inception of the field \cite{Spelke2007}: one view in which the mind is a relatively static assembly of special-purpose mechanisms developed by evolution, only capable of learning what is it programmed to acquire, and another view in which the mind is a general-purpose ``blank slate'' capable of turning arbitrary experience into knowledge and skills, and that could be directed at any problem.

A central point of this document is to make explicit and critically assess this dual definition that has been implicitly at the foundation of how we have been conceptualizing and evaluating intelligence in the context of AI research: crystallized skill on one hand, skill-acquisition ability on the other. Understanding this intellectual context and its ongoing influence is a necessary step before we can propose a formal definition of intelligence from a modern perspective.

\subsubsection{Intelligence as a collection of task-specific skills}
\label{intelligenceAsACollectionOfTaskSpecificSkills}

\epigraph{In the distant future I see open fields for far more important researches. Psychology will be based on a new foundation, that of the necessary acquirement of each mental power and capacity by gradation.}{Charles Darwin, 1859}

The evolutionary psychology view of human nature is that much of the human cognitive function is the result of special-purpose adaptations that arose to solve specific problems encountered by humans throughout their evolution (see e.g. \cite{CosmidesTooby1994, Pinker2002}) -- an idea which originated with Darwin \cite{Darwin1859} and that coalesced in the 1960s and 1970s. Around the same time that these ideas were gaining prominence in cognitive psychology, early AI researchers, perhaps seeing in electronic computers an analogue of the mind, mainly gravitated towards a view of intelligence as a set of static program-like routines, heavily relying on logical operators, and storing learned knowledge in a database-like memory.

This vision of the mind as a wide collection of vertical, relatively static programs that collectively implement ``intelligence'', was most prominently endorsed by influential AI pioneer Marvin Minsky (see e.g. \textit{The Society of Mind}, 1986 \cite{minsky1988society}). This view gave rise to definitions of intelligence and evaluation protocols for intelligence that are focused on task-specific performance. This is perhaps best illustrated by Minsky's 1968 definition of AI: \textit{``AI is the science of making machines capable of performing tasks that would require intelligence if done by humans''} \footnote{Note the lingering influence of the Turing Test.}. It was then widely accepted within the AI community that the ``problem of intelligence'' would be solved if only we could encode human skills into formal rules and encode human knowledge into explicit databases.

This view of intelligence was once so dominant that ``learning'' (discounted as pure memorization) was often not even mentioned at all in AI textbooks until the mid-1980s. Even McCarthy, a rare advocate for generality in AI, believed that the key to achieving generality was better knowledge bases \cite{mccarthy1987generality}. This definition and evaluation philosophy focused entirely on skill at narrow tasks normally handled by humans has led to a striking paradox, as pointed out by Hern{\'a}ndez-Orallo \cite{HernandezOrallo2016} in his 2017 survey: \textit{the field of artificial intelligence has been very successful in developing artificial systems that perform these tasks without featuring intelligence}, a trend that continues to this day.

\subsubsection{Intelligence as a general learning ability}
\label{intelligenceAsAGeneralLearningAbility}

\epigraph{Presumably the child brain is something like a notebook as one buys it from the stationer's. Rather little mechanism, and lots of blank sheets.}{Alan Turing, 1950}

In contrast, a number of researchers have taken the position that intelligence lies in the general ability to acquire new skills through learning; an ability that could be directed to a wide range of previously unknown problems -- perhaps even any problem at all. Contrast Minsky's task-focused definition of AI with the following one, paraphrased from McCarthy \cite{mccarthy1987generality} by  Hern{\'a}ndez-Orallo: \textit{``AI is the science and engineering of making machines do tasks they have never seen and have not been prepared for beforehand''} \cite{HernandezOrallo2016}.

The notion that machines could acquire new skills through a learning process similar to that of human children was initially laid out by Turing in his 1950 paper \cite{Turing1950}. In 1958, Friedberg noted astutely: \textit{``If we are ever to make a machine that will speak, understand or translate human languages, solve mathematical problems with imagination, practice a profession or direct an organization, either we must reduce these activities to a science so exact that we can tell a machine precisely how to go about doing them or we must develop a machine that can do things without being told precisely how''} \cite{friedberg1958learning}. But although the idea of generality through learning was given significant consideration at the birth of the field, and has long been championed by pioneers like McCarthy and Papert, it lay largely dormant until the resurgence of machine learning in the 1980s.

This view of intelligence echoes another long-standing conception of human nature that has had a profound influence on the history of cognitive science, contrasting with the evolutionary psychology perspective: Locke's \textit{Tabula Rasa} (blank slate), a vision of the mind as a flexible, adaptable, highly general process that turns experience into behavior, knowledge, and skills. This conception of the human mind can be traced back to Aristotle (\textit{De Anima}, c. 350BC, perhaps the first treatise of psychology \cite{Aristotle350BC}), was embraced and popularized by Enlightenment thinkers such as Hobbes \cite{hobbes}, Locke \cite{locke}, and Rousseau \cite{rousseau}. It has more recently found renewed vitality within cognitive psychology (e.g. \cite{RumelhartMcClelland1985}) and in AI via connectionism (e.g. \cite{hinton1993}).

With the resurgence of machine learning in the 1980s, its rise to intellectual dominance in the 2000s, and its peak as an intellectual quasi-monopoly in AI in the late 2010s via Deep Learning, a connectionist-inspired Tabula Rasa is increasingly becoming the dominant philosophical framework in which AI research is taking place. Many researchers are implicitly conceptualizing the mind via the metaphor of a ``randomly initialized neural network'' that starts blank and that derives its skills from ``training data'' --  a cognitive fallacy that echoes early AI researchers a few decades prior who conceptualized the mind as a kind of mainframe computer equipped with clever subroutines. We see the world through the lens of the tools we are most familiar with.\\

Today, it is increasingly apparent that both of these views of the nature of human intelligence -- either a collection of special-purpose programs or a general-purpose Tabula Rasa -- are likely incorrect, which we discuss in \ref{separatingTheInnateFromTheAcquired}, along with implications for artificial intelligence.

\subsection{AI evaluation: from measuring skills to measuring broad abilities}
\label{AIEvaluation}

These two conceptualizations of intelligence -- along with many other intermediate views combining elements from each side -- have influenced a host of approaches for evaluating intelligence in machines, in humans, and more rarely in both at the same time, which we discuss below. Note that this document is not meant as an extensive survey of AI evaluation methods -- for such a survey, we recommend Hern{\'a}ndez-Orallo 2017 \cite{MeasureOfAllMinds2017}. Other notable previous surveys include Cohen and Howe 1988 \cite{CohenHowe1988} and Legg and Hutter 2007 \cite{LeggHutter2007}.

\subsubsection{Skill-based, narrow AI evaluation}
\label{skillBasedNarrowAIEvaluation}

In apparent accordance with Minsky's goal for AI, the major successes of the field have been in building special-purpose systems capable of handling narrow, well-described tasks, sometimes at above human-level performance. This success has been driven by performance measures quantifying the skill of a system at a given task (e.g. how well an AI plays chess, how well an image classifier recognizes cats from dogs). There is no single, formalized way to do skill-based evaluation. Historically successful approaches include: 

\begin{itemize}
  \item Human review: having human judges observe the system's input-output response and score it. This is the idea behind the Turing test and its variants. This evaluation mode is rarely used in practice, due to being expensive, impossible to automate, and subjective. Some human-facing AI systems (in particular commercial chatbots) use it as one of multiple evaluation mechanics.
  
  \item White-box analysis: inspecting the implementation of the system to determine its input-output response and score it. This is most relevant for algorithms solving a fully-described task in a fully-described environment where all possible inputs can be explicitly enumerated or described analytically (e.g. an algorithm that solves the traveling salesman problem or that plays the game ``Connect Four''), and would often take the form of an optimality proof.
  
  \item Peer confrontation: having the system compete against either other AIs or humans. This is the preferred mode of evaluation for player-versus-player games, such as chess.
  
  \item Benchmarks: having the system produce outputs for a ``test set'' of inputs (or environments) for which the desired outcome is known, and score the response. 
\end{itemize}

Benchmarks in particular have been a major driver of progress in AI, because they are reproducible (the test set is fixed), fair (the test set is the same for everyone), scalable (it is inexpensive to run the evaluation many times), easy to set up, and flexible enough to be applicable to a wide range of possible tasks. Benchmarks have often been most impactful in the context of a competition between different research teams, such as the ILSVRC challenge for large-scale image recognition (ImageNet) \cite{imagenet} or the DARPA Grand Challenge for autonomous driving \cite{darpaGrandChallenge}. A number of private and community-led initiatives have been started on the premise that such benchmark-based competitions speed up progress (e.g. \href{https://kaggle.com}{Kaggle (kaggle.com)}, as well as academic alternatives such as \href{http://chalearn.org}{ChaLearn (chalearn.org)}, the \href{http://prize.hutter1.net/index.htm}{Hutter prize}, etc.), while some government organizations use competitions to deliberately trigger technological breakthroughs (e.g. DARPA, NIST).

These successes demonstrate the importance of \textit{setting clear goals and adopting objective measures of performance that are shared across the research community}. However, optimizing for a single metric or set of metrics often leads to tradeoffs and shortcuts when it comes to everything that isn't being measured and optimized for (a well-known effect on Kaggle, where winning models are often overly specialized for the specific benchmark they won and cannot be deployed on real-world versions of the underlying problem). In the case of AI, the focus on achieving task-specific performance while placing no conditions on \textit{how the system arrives at this performance} has led to systems that, despite performing the target tasks well, largely \textit{do not feature the sort of human intelligence that the field of AI set out to build}.

This has been interpreted by McCorduck as an ``AI effect'' where goalposts move every time progress in AI is made: \textit{``every time somebody figured out how to make a computer do something—play good checkers, solve simple but relatively informal problems—there was a chorus of critics to say, `that's not thinking' ''} \cite{McCorduck2004}. Similarly, Reed notes: \textit{``When we know how a machine does something `intelligent', it ceases to be regarded as intelligent. If I beat the world's chess champion, I'd be regarded as highly bright.''} \cite{Reed2006}. This interpretation arises from overly anthropocentric assumptions. As humans, we can only display high skill at a specific task if we have the ability to \textit{efficiently acquire skills in general}, which corresponds to intelligence as characterized in \ref{aNewPerspective}. No one is born knowing chess, or predisposed specifically for playing chess. Thus, if a human plays chess at a high level, we can safely assume that this person is intelligent, because we implicitly know that they had to use their \textit{general} intelligence to acquire this \textit{specific} skill over their lifetime, which reflects their general ability to acquire \textit{many other possible skills} in the same way. But the same assumption does not apply to a non-human system that does not arrive at competence the way humans do. If intelligence lies in the process of \textit{acquiring skills}, then \textit{there is no task X such that skill at X demonstrates intelligence}, unless X is actually a meta-task involving skill-acquisition across a broad range of tasks. The ``AI effect'' characterization is confusing the process of intelligence (such as the intelligence displayed by researchers creating a chess-playing program) with the artifact produced by this process (the resulting chess-playing program), due to these two concepts being fundamentally intertwined in the case of humans. We discuss this further in \ref{criticalAssessment}.

Task-specific performance is a perfectly appropriate and effective measure of success if and only if handling the task as initially specified is the end goal of the system -- in other words, if our measure of performance captures exactly what we expect of the system. However, it is deficient if we need systems that can show autonomy in handling situations that the system creator did not plan for, that can dynamically adapt to changes in the task -- or in the context of the task -- without further human intervention, or that can be repurposed for other tasks. Meanwhile, robustness and flexibility are increasingly being perceived as important requirements for certain broader subfields of AI, such as L5 self-driving, domestic robotics, or personal assistants; there is even increasing interest in generality itself (e.g. developmental robotics \cite{asada2009}, artificial general intelligence \cite{GoertzelPennachin2007}). This points to a need to move beyond skill-based evaluation for such endeavours, and to find ways to evaluate robustness and flexibility, especially in a cross-task setting, up to generality. But what do we really mean when we talk about robustness, flexibility, and generality?

\subsubsection{The spectrum of generalization: robustness, flexibility, generality}
\label{spectrumOfGeneralization}

\epigraph{Even though such machines might do some things as well as we do them, or perhaps even better, they would inevitably fail in others, which would reveal they were acting not through understanding, but only from the disposition of their organs.}{Ren\'{e} Descartes, 1637}

The resurgence of machine learning in the 1980s has led to an interest in formally defining, measuring, and maximizing \textit{generalization}. Generalization is a concept that predates machine learning, originally developed to characterize how well a statistical model performs on inputs that were not part of its training data. In recent years, the success of Deep Learning \cite{lecun2015deep}, as well as increasingly frequent run-ins with its limitations (see e.g. \cite{LakeUTG16, chollet2017, marcus2018deep}), have triggered renewed interest in generalization theory in the context of machine learning (see e.g. \cite{rethinkingGeneralization, neyshabur2017exploring, Jiang2018PredictingTG, Packer2018AssessingGI, CoinRun, Juliani2019}). The notion of generalization can be formally defined in various contexts (in particular, statistical learning theory \cite{Vapnik1995} provides a widely-used formal definition that is relevant for machine learning, and we provide a more general formalization in \ref{aFormalSynthesis}). We can informally define ``generalization'' or ``generalization power'' for any AI system to broadly mean \textit{``the ability to handle situations (or tasks) that differ from previously encountered situations''}.\\

The notion of ``previously encountered situation'' is somewhat ambiguous, so we should distinguish between two types of generalization:

\begin{itemize}
  \item \textbf{System-centric generalization}: this is the ability of a learning system to handle situations it has not itself encountered before. The formal notion of generalization error in statistical learning theory would belong here.
    \begin{itemize}
     \item For instance, if an engineer develops a machine learning classification algorithm and fits it on a training set of $N$ samples, the ``generalization'' of this learning algorithm would refer to its classification error over images not part of the training set.
    \item Note that the generalization power of this algorithm may be in part due to \textit{prior knowledge} injected by the developer of the system. This prior knowledge is ignored by this measure of generalization.
    \end{itemize}
 \item  \textbf{Developer-aware generalization}: this is the ability of a system, either learning or static, to handle situations that neither the system nor the developer of the system have encountered before.
    \begin{itemize}
    \item For instance, if an engineer uses a ``development set'' of $N$ samples to create a static classification algorithm that uses hard-coded heuristic rules, the ``generalization'' of this static algorithm would refer to its classification error over images not part of the ``development set''.
    \item Note that ``developer-aware generalization'' is equivalent to ``system-centric generalization'' if we include the developer of the system as part of the system.
    \item Note that ``developer-aware generalization'' accounts for any prior knowledge that the developer of the system has injected into it. ``System-centric generalization'' does not.
    \end{itemize}
\end{itemize}

In addition, we find it useful to qualitatively define \textit{degrees of generalization} for information-processing systems:

\begin{itemize}
    \item \textbf{Absence of generalization}: The notion of generalization as we have informally defined above fundamentally relies on the related notions of novelty and uncertainty: a system can only generalize to novel information that could not be known in advance to either the system or its creator. AI systems in which there is no uncertainty do not display generalization. For instance, a program that plays tic-tac-toe via exhaustive iteration cannot be said to ``generalize'' to all board configurations. Likewise, a sorting algorithm that is proven to be correct cannot be said to ``generalize'' to all lists of integers, much like proven mathematical statements cannot be said to ``generalize'' to all objects that match the assumptions of their proof \footnote{This is a distinct definition from ``generalization'' in mathematics, where ``to generalize'' means to extend the scope of application of a statement by weakening its assumptions.}.
    
    \item \textbf{Local generalization, or ``robustness''}: This is the ability of a system to handle new points from a known distribution for a single task or a well-scoped set of known tasks, given a sufficiently dense sampling of examples from the distribution (e.g. tolerance to anticipated perturbations within a fixed context). For instance, an image classifier that can distinguish previously unseen 150x150 RGB images containing cats from those containing dogs, after being trained on many such labeled images, can be said to perform local generalization. One could characterize it as \textit{``adaptation to known unknowns within a single task or well-defined set of tasks''}. This is the form of generalization that machine learning has been concerned with from the 1950s up to this day.

    \item \textbf{Broad generalization, or ``flexibility''}: This is the ability of a system to handle a broad category of tasks and environments without further human intervention. This includes the ability to handle situations that could not have been foreseen by the creators of the system. This could be considered to reflect human-level ability in a single broad activity domain (e.g. household tasks, driving in the real world), and could be characterized as \textit{``adaptation to unknown unknowns across a broad category of related tasks''}. For instance, a L5 self-driving vehicle, or a domestic robot capable of passing Wozniak's coffee cup test (entering a random kitchen and making a cup of coffee) \cite{Wozniak2007} could be said to display broad generalization. Arguably, even the most advanced AI systems today do not belong in this category, although there is increasing research interest in achieving this level.
    
    \item \textbf{Extreme generalization}: This describes open-ended systems with the ability to handle entirely new tasks that only share abstract commonalities with previously encountered situations, applicable to any task and domain within a wide scope. This could be characterized as ``adaptation to unknown unknowns across an unknown range of tasks and domains''. Biological forms of intelligence (humans and possibly other intelligent species) are the only example of such a system at this time. A version of extreme generalization that is of particular interest to us throughout this document is \textit{human-centric extreme generalization}, which is the specific case where the scope considered is the space of tasks and domains that fit within the human experience. We will refer to ``human-centric extreme generalization'' as ``generality''. Importantly, as we deliberately define generality here by using human cognition as a reference frame (which we discuss in \ref{theMeaningOfGenerality}), it is only ``general'' in a limited sense. Do note, however, that humans display extreme generalization both in terms of system-centric generalization (quick adaptability to highly novel situations from little experience) and developer-aware generalization (ability of contemporary humans to handle situations that previous humans have never experienced during their evolutionary history).

\end{itemize}

To this list, we could, theoretically, add one more entry: \textit{``universality''}, which would extend ``generality'' beyond the scope of task domains relevant to humans, to any task that could be practically tackled within our universe (note that this is different from ``any task at all'' as understood in the assumptions of the No Free Lunch theorem \cite{Wolpert1997, wolpert2012no}). We discuss in \ref{theMeaningOfGenerality} why we do not consider universality to be a reasonable goal for AI.

Crucially, the history of AI has been one of slowly climbing up this spectrum, starting with systems that largely did not display generalization (symbolic AI), and evolving towards robust systems (machine learning) capable of local generalization. We are now entering a new stage, where we seek to create flexible systems capable of broad generalization (e.g. hybrid symbolic and machine learning systems such as self-driving vehicles, AI assistants, or cognitive developmental robots). Skill-focused task-specific evaluation has been appropriate for close-ended systems that aim at robustness in environments that only feature known unknowns, but developing systems that are capable of handling unknown unknowns requires evaluating their \textit{abilities} in a general sense.

\begin{figure}[h]
    \centering
    \includegraphics[scale=0.35]{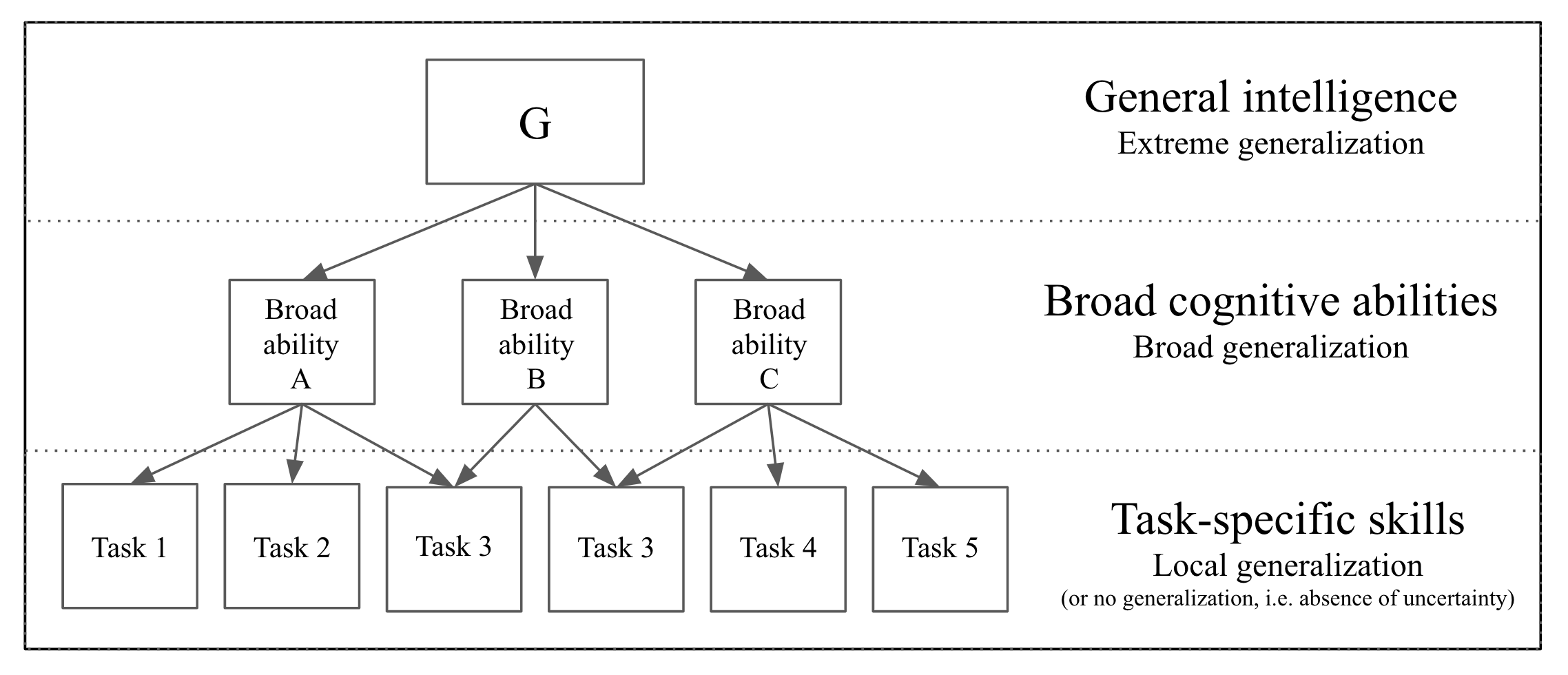}
    \caption{Hierarchical model of cognitive abilities and its mapping to the spectrum of generalization.}
    \label{fig:three-strata}
\end{figure}

Importantly, the spectrum of generalization outlined above seems to mirror the organization of humans cognitive abilities as laid out by theories of the structure of intelligence in cognitive psychology. Major theories of the structure of human intelligence (CHC \cite{CHC}, g-VPR \cite{gVPR}) all organize cognitive abilities in a hierarchical fashion (figure \ref{fig:three-strata}), with three strata (in CHC): general intelligence (g factor) at the top, broad abilities in the middle, and specialized skills or test tasks at the bottom (this extends to 4 strata for g-VPR, which splits broad abilities into two layers), albeit the taxonomy of abilities differs between theories. Here, ``extreme generalization'' corresponds to the g factor, ``broad generalization'' across a given domain corresponds to a broad cognitive ability, and ``local generalization'' (as well as the no-generalization case) corresponds to task-specific skill.

Measuring such broad abilities (and possibly generality itself) rather than specific skills has historically been the problematic of the field of psychometrics. Could psychometrics inform the evaluation of abilities in AI systems?

\noindent Note that, in what follows:

\begin{itemize}
    \item We use ``broad abilities'' to refer to cognitive abilities that lead to broad or extreme generalization. Developing such abilities should be the goal of any researcher interested in flexible AI or general AI. ``Broad abilities'' is often meant in opposition to ``local generalization''.
    \item We use ``generalization'' to refer to the entire spectrum of generalization, starting with local generalization.
    \item Because human general intelligence (the g factor) is itself a very broad cognitive ability (the top of the hierarchy of abilities), we use the term ``intelligence'' or ``general intelligence'' to refer to extreme generalization as defined above.

\end{itemize}

\subsubsection{Measuring broad abilities and general intelligence: the psychometrics perspective}
\label{measuringBroadAbilities}

\epigraph{It seems to us that in intelligence there is a fundamental faculty, the alteration or the lack of which, is of the utmost importance for practical life. This faculty is [...] the faculty of adapting one's self to circumstances.}{Alfred Binet, 1916}

In the early days of the 20th century, Binet and Simon, looking for a formal way to distinguish children with mental disabilities from those with behavior problems, developed the Binet-Simon scale \cite{BinetSimon1904}, the first test of intelligence, founding the field of psychometrics. Immediately after, Spearman observed that individual results across different, seemingly unrelated types of intelligence tests were correlated, and hypothesized the existence of a single factor of general intelligence, the g factor \cite{Spearman1904, Spearman1927}. Today, psychometrics is a well-established subfield of psychology that has arrived at some of the most reproducible results of the field. Modern intelligence tests are developed by following strict standards regarding reliability (low measurement error, a notion tied to reproducibility), validity (measuring what one purports to be measuring, a notion tied to statistical consistency and predictiveness), standardization, and freedom from bias -- see e.g. Classical Test Theory (CTT) \cite{CTT} and Item Response Theory (IRT) \cite{IRT}.

A fundamental notion in psychometrics is that intelligence tests evaluate broad cognitive \textit{abilities} as opposed to task-specific \textit{skills}. Theories of the structure of intelligence (such as CHC, g-VPR), which have co-evolved with psychometric testing (statistical phenomena emerging from test results have informed these theories, and these theories have informed test design) organize these abilities in a hierarchical fashion (figure \ref{fig:three-strata}), rather similarly to the spectrum of generalization we presented earlier. Importantly, an \textit{ability} is an abstract construct (based on theory and statistical phenomena) as opposed to a directly measurable, objective property of an individual mind, such as a score on a specific test. Broad abilities in AI, which are also constructs, fall into the exact same evaluation problematics as cognitive abilities from psychometrics. Psychometrics approaches the quantification of abilities by using broad batteries of test tasks rather than any single task, and by analysing test results via probabilistic models. Importantly, the tasks should be previously unknown to the test-taker, i.e., we assume that test-takers do not practice for intelligence tests. This approach is highly relevant to AI evaluation.

Remarkably, in a parallel to psychometrics, there has been recent and increasing interest across the field of AI in using broad batteries of test tasks to evaluate systems that aim at greater flexibility. Examples include the Arcade Learning Environment for Reinforcement Learning agents \cite{Bellemare2013}, Project Malm{\"O} \cite{Malmo}, the Behavior Suite \cite{osband2019behaviour}, or the GLUE \cite{wang2018glue} and SuperGLUE \cite{wang2019superglue} benchmarks for natural language processing. The underlying logic of these efforts is to measure something more general than skill at one specific task by broadening the set of target tasks. However, when it comes to assessing flexibility, a critical defect of these multi-task benchmarks is that the set of tasks is still known in advance to the developers of any test-taking system, and it is fully expected that test-taking systems will be able to practice specifically for the target tasks, leverage task-specific built-in prior knowledge inherited from the system developers, leverage external knowledge obtained via pre-training, etc. As such, these benchmarks still appear to be highly gameable (see e.g. \ref{measuringTheRightThing}) -- merely widening task-specific skill evaluation to more tasks does not produce a qualitatively different kind of evaluation. Such benchmarks are still looking at skills, rather than abilities, in contrast with the psychometrics approach (this is not to say that such benchmarks are not useful; merely that such static multi-task benchmarks do not directly assess flexibility or generality).

In addition to these multi-task benchmarks, a number of more ambitious test suites for cognitive abilities of AI have been proposed in the past but have not been implemented in practice: the Newell test by Anderson and Lebiere (\cite{NewellTest}, named in reference to \cite{Newell1973}), the BICA ``cognitive decathlon'' targeted at developmental robotics \cite{BICA}, the Turing Olympics \cite{turingOlympics}, and the I-Athlon \cite{adams2016athlon}. Lacking concrete implementations, it is difficult to assess whether these projects would have been able to address the ability evaluation problem they set out to solve. On the other hand, two similarly-spirited but more mature test suite have emerged recently, focused on generalization capabilities as opposed to specific tasks: the Animal-AI Olympics \cite{beyret2019animalai} (\href{http://animalaiolympics.com/}{animalaiolympics.com}) and the GVGAI competition \cite{perez2018general} (\href{http://www.gvgai.net/}{gvgai.net}). Both take the position that AI agents should be evaluated on an unseen set of tasks or games, in order to test learning or planning abilities rather than special-purpose skill. Both feature a multi-game environment and an ongoing public competition.

\subsubsection{Integrating AI evaluation and psychometrics}
\label{integrationAIEvaluationAndPsychometric}

Besides efforts to broaden task-specific evaluation to batteries of multi-task tests, there have been more direct and explicit attempts to integrate AI evaluation and psychometrics. A first approach is to reuse existing psychometric intelligence tests, initially developed for humans, as a way to assess intelligence in AI systems -- perhaps an obvious idea if we are to take the term ``artificial intelligence'' literally. This idea was first proposed by Green in 1964 \cite{green1964intelligence}, and was, around the same time, explored by Evans \cite{evans1968}, who wrote a LISP program called ANALOGY capable of solving a geometric analogy task of the kind that may be found in a pyschometric intelligence test. Newell suggested the idea again in 1973 \cite{Newell1973} in his seminal paper \textit{You can't play 20 questions with Nature and win}. It was proposed again and refined by Bringsjord et al. in the 2000s under the name ``Psychometric AI'' (PAI) \cite{Bringsjord2003}. However, it has since become apparent that it is possible for AI system developers to game human intelligence tests, because the tasks used in these tests are available to the system developers, and thus the developers can straightforwardly solve the abstract form of these problems themselves and hard-code the solution in program form (see, for instance, \cite{Detterman2011, SanghiDowe2003, HODowe2012}), much like Evans did with in the 1960s with the ANALOGY program. Effectively, in this case, it is the system developers who are solving the test problems, rather than any AI. The implicit assumptions that psychometric test designers make about human test-takers turn out to be difficult to enforce in the case of machines.

An alternative, more promising approach is to leverage what psychometrics can teach us about ability assessment and test design to create new types of benchmarks targeted specifically at evaluating broad abilities in AI systems. Along these lines,  Hern{\'a}ndez-Orallo et al. have proposed extending psychometric evaluation to any intelligent system, including AI agents and animals, in ``Universal Psychometrics'' \cite{HernandezOrallo2014}.

We argue that several important principles of psychometrics can inform intelligence evaluation in AI in the context of the development of broad AI and general AI:

\begin{itemize}
    \item Measuring abilities (representative of broad generalization and skill-acquisition efficiency), not skills. Abilities are distinct from skills in that they induce broad generalization, i.e. they form the basis for skill across a broad range of tasks, including tasks that were previously unknown to the ability-enabled system and its developers.
    \item Doing so via batteries of tasks rather than any single task, \textit{that should be previously unknown to both the test taking system and the system developers} (this is necessary to assess broad generalization as opposed to skill or local generalization).
    \item Having explicit standards regarding \textit{reliability, validity, standardization}, and \textit{freedom from bias}:
        \begin{itemize}
            \item Reliability implies that the test results for a given system should be reproducible over time and across research groups.
            \item Validity implies that what the test assesses should be clearly understood; test creators should be able to answer 1) what assumptions does the test make? 2) what does the test predict, i.e. what broad abilities would a successful result demonstrate, and how well does the test predict these abilities? (Which should ideally be achieved via statistical quantification.)
            \item Standardization implies adopting shared benchmarks across the subset of the research community that pursues broad AI and general AI. Standard benchmarks in computer vision and natural language processing have already shown to be highly effective catalyzers of progress.
            \item Freedom from bias implies that the test should not be biased against groups of test-takers in ways that run orthogonal to the abilities being assessed. For instance, a test of intelligence designed for both humans and AI should not leverage uniquely human acquired knowledge, or should not involve constraints unrelated to intelligence within which machines have unfair advantages (such as fast reaction times), etc.
        \end{itemize}
\end{itemize}

Simultaneously, we argue that certain other aspects of psychometrics may be discarded in the development of new intelligence tests for AI:

\begin{itemize}
    \item The exact number and taxonomy of cognitive abilities considered, being a subject of ongoing debate within cognitive psychology and being perhaps overly anthropocentric, should not be used as a strict template for artificial cognitive architectures and their evaluation. Existing taxonomies may at best serve as a source of inspiration.
    \item A number of abilities being assessed by psychometric intelligence tests are crystallized abilities (e.g. reading and writing), i.e. abilities that are acquired through experience, which are not clearly distinguishable from skills (they are effectively multi-purpose skills). We argue that AI tests that seek to assess flexibility and generality should not consider crystallized abilities, but rather, should focus on abilities that enable \textit{new skill acquisition}. If a system possesses abilities that enable efficient skill-acquisition in a domain, the system should have no issue in developing corresponding skills and crystallized abilities.

\end{itemize}

\subsubsection{Current trends in broad AI evaluation}
\label{currentTrendsInAIEvaluation}

Despite a rising interest in building flexible systems, or even in generality itself, for the most part the AI community has not been paying much attention to psychometric evaluation, Psychometric AI, or Universal Psychometrics. If we are to assess the contemporary zeitgeist of broad AI evaluation, here is what we see. \footnote{Because broad AI research is currently largely dominated by Reinforcement Learning (RL) approaches, many of our observations here are specific to RL.}

First, we note several positive developments. Since 2017, there is increasing awareness that one should seek to establish some form of generalization in evaluating Reinforcement Learning (RL) algorithms (e.g. \cite{justesen2018illuminating, Packer2018AssessingGI, CoinRun, Juliani2019}), which was previously a stark problem \cite{Rajeswaran2017, Henderson2018, zhang2018dissection, Packer2018AssessingGI}, as RL agents have for a long time been tested on their training data. Further, there is increasing interest in evaluating the data-efficiency of learning algorithms (e.g. \cite{buckman2018sampleefficient}), in particular in the context of RL for games such as Atari games or Minecraft (e.g. \cite{Malmo, MineRL}). Lastly, as noted in \ref{measuringBroadAbilities}, there has been a trend towards leveraging multi-task benchmarks as a way to assess robustness and flexibility (e.g. \cite{Bellemare2013, Malmo, osband2019behaviour, wang2018glue, wang2019superglue}).

Unfortunately, we must also note several negatives. The robustness of the systems being developed, in particular Deep Learning models, is often problematic (see e.g. \cite{chollet2017, marcus2018deep}). This is due in large part to the fact that most benchmarks do not pay much attention to formally assessing robustness and quantifying generalization, and thus can be solved via ``shortcuts'' that gradient descent is apt at exploiting (e.g. surface statistics such as textures in the case of computer vision \cite{Jo2017MeasuringTT}). Likewise, the reproducibility (reliability) of research findings is often an issue \cite{Pineau2018}, especially in Reinforcement Learning, although some progress has been made on this front.

Most importantly, the evaluation of any ability that goes decisively beyond local generalization is still largely a green field, and little effort has been devoted to investigate it. Hern{\'a}ndez-Orallo noted in 2017 that \textit{``ability-oriented and general-purpose evaluation approaches [...] are still very incipient, and more research and discussion is needed''} \cite{HernandezOrallo2016}. Recent attempts at broadening task-specific benchmarks by including multiple tasks do not measure developer-aware generalization, as the tasks are all known in advance to system developers (as noted in \ref{measuringBroadAbilities}). Attempts at assessing generalization by testing RL systems on previously unseen game levels, like CoinRun \cite{CoinRun} or Obstacle Tower \cite{Juliani2019}, are still only looking at task-specific local generalization, by evaluating a candidate system on new samples from a known distribution rather than using a substantially new task (as suggested in \ref{ARCPossibleAlternatives}). In addition, the fact the level-generation programs used are available to the AI developers means it is possible to ``cheat'' on these benchmarks by sampling arbitrary amounts of training data (cf. \ref{measuringTheRightThing}).

Further, contemporary research ``moonshots'' that are publicly advertised as being steps towards general intelligence appear to still be focusing on skill-based task-specific evaluation for board games and video games (e.g. Go \cite{silver2017mastering, silver2017mastering2} and \href{https://starcraft.com}{StarCraft} \cite{Vinyals2017StarCraftIA} for DeepMind, \href{https://dota2.com}{DotA2} \cite{OpenAIDota2} for OpenAI) via highly-mediatized confrontations with top human players. Despite claims of progress towards general AI in associated public communications\footnote{OpenAI public statement: \textit{``Five is a step towards advanced AI systems which can handle the complexity and uncertainty of the real world''}}, such evaluation does not involve any measure of generalization power, and has little-to-no overlap with the development of flexibility and generality, as we outline in \ref{criticalAssessment}. For example, although OpenAI's DotA2-playing AI ``Five'' was trained on 45,000 years of play and was able to beat top human players \cite{OpenAIDota2}, it has proven very brittle, as non-champion human players were able to find strategies to reliably beat it in a matter of days after the AI was made available for the public to play against \cite{OpenAIFiveArenaResults}. In addition, Five did not even generalize to DotA2 in the first place: it could only play a restricted version of the game, with 16 characters instead of over 100. Likewise, AlphaGo and its successor AlphaZero, developed in 2016 and 2017, have not yet found any application outside of board games, to the best of our knowledge.\\

We deplore this discrepancy between a focus on surpassing humans at tests of skill on one hand (while entirely disregarding whether the methods through which skill is achieved are generalizable), and a manifest interest in developing broad abilities on the other hand -- an endeavour entirely orthogonal to skill itself. We hypothesize that this discrepancy is due to a lack of a clear conceptualization of intelligence, skill, and generalization, as well as a lack of appropriate measures and benchmarks for broad cognitive abilities. In what follows, we expose in more detail the issue with using task-specific ``moonshots'' (e.g. achieving better-than-human performance in a video game or board game) as stepping stones towards more general forms of AI, and we propose a formal definition of intelligence meant to be actionable in the pursuit of flexible AI and general AI.


\section{A new perspective}
\label{aNewPerspective}

\subsection{Critical assessment}
\label{criticalAssessment}

\subsubsection{Measuring the right thing: evaluating skill alone does not move us forward}
\label{measuringTheRightThing}

In 1973, psychologist and computer science pioneer Allen Newell, worried that recent advances in cognitive psychology were not bringing the field any closer to a holistic theory of cognition, published his seminal paper \textit{You can't play 20 questions with nature and win} \cite{Newell1973}, which helped focus research efforts on cognitive architecture modelling, and provided new impetus to the longstanding quest to build a chess-playing AI that would outperform any human. Twenty-four years later, In 1997, IBM's DeepBlue beat Gary Kasparov, the best chess player in the world, bringing this quest to an end \cite{Campbell2002}. When the dust settled, researchers were left with the realization that building an artificial chess champion had not actually taught them much, if anything, about human cognition. They had learned how to build a chess-playing AI, and neither this knowledge nor the AI they had built could generalize to anything other than similar board games.

It may be obvious from a modern perspective that a static chess-playing program based on minimax and tree search would not be informative about human intelligence, nor competitive with humans in anything other than chess. But it was not obvious in the 1970s, when chess-playing was thought by many to capture, and require, the entire scope of rational human thought. Perhaps less obvious in 2019 is that efforts to ``solve'' complex video games using modern machine learning methods still follow the same pattern. Newell wrote \cite{Newell1973}: \textit{``we know already from existing work [psychological studies on humans] that the task [chess] involves forms of reasoning and search and complex perceptual and memorial processes. For more general considerations we know that it also involves planning, evaluation, means-ends analysis and redefinition of the situation, as well as several varieties of learning -- short-term, post-hoc analysis, preparatory analysis, study from books, etc.''}. The assumption was that solving chess would require implementing these general abilities. Chess does indeed involve these abilities -- in humans. But while possessing these general abilities makes it possible to solve chess (and many more problems), by going from the general to the specific, inversely, there is no clear path from the specific to the general. Chess does not \textit{require} any of these abilities, and can be solved by taking radical shortcuts that run orthogonal to human cognition.

Optimizing for single-purpose performance is useful and valid if one's measure of success can capture exactly what one seeks (as we outlined in \ref{skillBasedNarrowAIEvaluation}), e.g. if one's end goal is a chess-playing machine and nothing more. But from the moment the objective is settled, the process of developing a solution will be prone to taking all shortcuts available to satisfy the objective of choice -- whether this process is gradient descent or human-driven research. These shortcuts often come with undesirable side-effects when it comes to considerations not incorporated the measure of performance. If the environment in which the system is to operate is too unpredictable for an all-encompassing objective function to be defined beforehand (e.g. most real-world applications of robotics, where systems face unknown unknowns), or if one aims at a general-purpose AI that could be applied to a wide range of problems with no or little human engineering, then one must somehow \textit{optimize directly for flexibility and generality}, rather than solely for performance on any specific task.

This is, perhaps, a widely-accepted view today when it comes to static programs that hard-code a human-designed solution. When a human engineer implements a chatbot by specifying answers for each possible query via if/else statements, we do not assume this chatbot to be intelligent, and we do not expect it to generalize beyond the engineer's specifications. Likewise, if an engineer looks at a specific IQ test task, comes up with a solution, and write down this solution in program form, we do not expect the program to generalize to new tasks, and we do not believe that the program displays intelligence -- the only intelligence at work here is the engineer's. The program merely encodes the crystallized output of the engineer's thought process -- it is this process, not its output, that implements intelligence. Intelligence is not demonstrated by the performance of the output program (a skill), but by the fact that the same process can be applied to a vast range of previously unknown problems (a general-purpose ability): the engineer's mind is capable of extreme generalization. Since the resulting program is merely encoding the output of that process, it is no more intelligent than the ink and paper used to write down the proof of a theorem.

However, what of a program that is not hard-coded by humans, but trained from data to perform a task? A learning machine certainly \textit{may} be intelligent: learning is a necessary condition to adapt to new information and acquire new skills. But being programmed through exposure to data is no guarantee of generalization or intelligence. Hard-coding prior knowledge into an AI  is not the only way to artificially ``buy'' performance on the target task without inducing any generalization power. There is another way: adding more training data, which can augment skill in a specific vertical or task without affecting generalization whatsoever.

Information processing systems form a spectrum between two extremes: on one end, static systems that consist entirely of hard-coded priors (such as DeepBlue or our if/else chatbot example), and on the opposite end, systems that incorporate very few priors and are almost entirely programmed via exposure to data (such as a hashtable or a densely-connected neural network). Most intelligent systems, including humans and animals, combine ample amounts of both priors and experience, as we point out in \ref{separatingTheInnateFromTheAcquired}. Crucially, the ability to generalize is an axis that runs orthogonal to the prior/experience plane. Given a learning system capable of achieving a certain level of generalization, modifying the system by incorporating more priors or more training data about the task can lead to greater task-specific performance without affecting generalization. In this case, both priors and experience serve as a way to ``game'' any given test of skill without having to display the sort of general-purpose abilities that humans would rely on to acquire the same skill.

This can be readily demonstrated with a simple example: consider a hashtable that uses a locality-sensitive hash function (e.g. nearest neighbor) to map new inputs to previously seen inputs. Such a system implements a learning algorithm capable of local generalization, the extent of which is fixed (independent of the amount of data seen), determined only by the abstraction capabilities of the hash function. This system, despite only featuring trace amounts of generalization power, is already sufficient to ``solve'' any task for which unlimited training data can be generated, such as any video game. All that one has to do is obtain a dense sampling of the space of situations that needs to be covered, and associate each situation with an appropriate action vector.

Adding ever more data to a local-generalization learning system is certainly a fair strategy if one's end goal is skill on the task considered, but it will not lead to generalization beyond the data the system has seen (the resulting system is still very brittle, e.g. Deep Learning models such as OpenAI Five), and crucially, developing such systems \textit{does not teach us anything about achieving flexibility and generality}. ``Solving'' any given task with beyond-human level performance by leveraging either unlimited priors or unlimited data does not bring us any closer to broad AI or  general AI, whether the task is chess, football, or any e-sport.

Current evidence (e.g. \cite{LakeUTG16, Jo2017MeasuringTT, chollet2017, marcus2018deep, justesen2018illuminating}) points to the fact that contemporary Deep Learning models are local-generalization systems, conceptually similar to a locality-sensitive hashtable -- they may be trained to achieve arbitrary levels of skill at any task, but doing so requires a dense sampling of the input-cross-target space considered (as outlined in \cite{chollet2017}), which is impractical to obtain for high-value real-world applications, such as L5 self-driving (e.g. \cite{bansal2018chauffeurnet} notes that 30 million training situations is not enough for a Deep Learning model to learn to drive a car in a plain supervised setting). Hypothetically, it may be shown in the future that methods derived from Deep Learning could be capable of stronger forms of generalization, but demonstrating this cannot be done merely by achieving high skill, such as beating humans at DotA2 or Starcraft given unlimited data or unlimited engineering; instead, one should seek to precisely establish and quantify the \textit{generalization strength} of such systems (e.g. by considering prior-efficiency and data-efficiency in skill acquisition, as well as the developer-aware generalization difficulty of the tasks considered). A central point of this document is to provide a formal framework for doing so (\ref{aFormalSynthesis} and \ref{evaluatingIntelligenceInThisLight}). Failing to account for priors, experience, and generalization difficulty in our evaluation methods will prevent our field from climbing higher along the spectrum of generalization (\ref{spectrumOfGeneralization}) and from eventually reaching general AI.\\

In summary, the hallmark of broad abilities (including general intelligence, as per \ref{theMeaningOfGenerality}) is the power to adapt to change, acquire skills, and solve previously unseen problems -- not skill itself, which is merely the crystallized output of the process of intelligence. Testing for skill at a task that is known in advance to system developers (as is the current trend in general AI research) can be gamed without displaying intelligence, in two ways: 1) unlimited prior knowledge, 2) unlimited training data. To actually assess broad abilities, and thus make progress toward flexible AI and eventually general AI, it is imperative that we control for \textit{priors}, \textit{experience}, and \textit{generalization difficulty} in our evaluation methods, in a rigorous and quantitative way.

\subsubsection{The meaning of generality: grounding the g factor}
\label{theMeaningOfGenerality}

It is a well-known fact of cognitive psychology that different individuals demonstrate different cognitive abilities to varying degrees, albeit results across all tests of intelligence are correlated. This points to cognition being a multi-dimensional object, structured in a hierarchical fashion (figure \ref{fig:three-strata}), with a single generality factor at the top, the g factor. But is ``general intelligence'' the apex of the cognitive pyramid in an absolute sense (as is sometimes assumed by proponents of ``Artificial General Intelligence''), or is it merely a broader cognitive ability, one that would remain fairly specialized, and wouldn't be qualitatively distinct from other abilities lower down the hierarchy? How general is human intelligence?

The No Free Lunch theorem \cite{Wolpert1997, wolpert2012no} teaches us that any two optimization algorithms (including human intelligence) are equivalent when their performance is averaged across \textit{every possible problem}, i.e. algorithms should be
tailored to their target problem in order to achieve better-than-random performance. However, what is meant in this context by ``every possible problem'' refers to a uniform distribution over problem space; the distribution of tasks that would be practically relevant to our universe (which, due to its choice of laws of physics, is a specialized environment) would not fit this definition. Thus we may ask: is the human g factor universal? Would it generalize to every possible task in the universe?

 This is a question that is largely irrelevant for psychometrics, because as a subfield of psychology, it makes the implicit assumption that it is concerned solely with humans and the human experience. But this question is highly relevant when it comes to AI: if there is such a thing as universal intelligence, and if human intelligence is an implementation of it, then this algorithm of universal intelligence should be the end goal of our field, and reverse-engineering the human brain could be the shortest path to reach it. It would make our field close-ended: a riddle to be solved. If, on the other hand, human intelligence is a broad but ad-hoc cognitive ability that generalizes to human-relevant tasks but not much else, this implies that AI is an open-ended, fundamentally anthropocentric pursuit, tied to a specific scope of applicability. This has implications for how we should measure it (by using human intelligence and human tasks as a reference) and for the research strategies we should follow to achieve it.
 
The g factor, by definition, represents the single cognitive ability common to success across all intelligence tests, emerging from applying factor analysis to test results across a diversity of tests and individuals. But intelligence tests, by construction, only encompass tasks that humans can perform -- tasks that are immediately recognizable and understandable by humans (anthropocentric bias), since including tasks that humans couldn't perform would be pointless. Further, psychometrics establishes measurement validity by demonstrating predictiveness with regard to activities that humans value (e.g. scholastic success): the very idea of a ``valid'' measure of intelligence only makes sense within the frame of reference of human values.

In fact, the interpretation of what specific abilities make someone ``intelligent'' vary from culture to culture \cite{yang1997taiwanese, Sternberg2004, cocodia2014cultural}. More broadly, humans have historically had a poor track record when it comes to attributing intelligence to complex information-processing agents around them, whether looking at humans from other cultures or at animals (such as octopuses, dolphins, great apes, etc.). We only reluctantly open up to the possibility that systems different from ourselves may be ``intelligent'' if they display relatable human-like behaviors that we associate with intelligence, such as language or tool use; behaviors that have high intrinsic complexity and high adaptability but that are not directly relatable (such as octopus camouflage) are not perceived as intelligent. This observation extends to collective entities (e.g. markets, companies, Science as an institution) and natural processes (e.g. biological evolution). Although they can be modeled as standalone systems whose abilities and behavior match broadly accepted definitions of intelligence (achieving goals across a wide range of environments, demonstrating flexibility and adaptability, etc.), we do not categorize these systems as intelligent, simply because they aren't sufficiently human-like.

To use a well-known cross-domain analogy \cite{flynn2007intelligence}: much like ``intelligence'', the notion of ``physical fitness'' (as it pertains to sports and other physical activities) is an intuitively-understandable, informal, yet useful concept. Like intelligence, fitness is not easily reducible to any single factor (such as a person's age or muscle mass), rather, it seems to emerge from a constellation of interdependent factors. If we sought to rigorously measure physical fitness in humans, we would come up with a set of diverse tests such as running a 100m, running a marathon, swimming, doing sit-ups, doing basketball throws, etc., not unlike IQ test suites. Across tests results, we would observe clusters of correlations, corresponding to broad ``physical abilities'' strictly analogous to cognitive abilities (e.g. lung capacity might be such an ``ability'' inducing correlations across tests). Much like in the case of cognitive abilities, experts would probably disagree and debate as to the exact taxonomy of these broad abilities (is being ``tall and lean'' an ability, or is ``tallness'' a standalone factor?). And crucially, we should intuitively expect to find that \textit{all tests results would be correlated}: we would observe a physical g factor, corresponding to the general intuitive construct of ``physical fitness''.

But would this mean that human morphology and motor affordances are ``general'' in an absolute sense, and that a very fit person could handle any physical task at all? Certainly not; we are not adapted for the large majority of environments that can be found in the universe -- from the Earth's oceans to the surface of Venus, from the atmosphere of Jupiter to interstellar space. It is, however, striking and remarkable that human physical abilities generalize to a far greater range of environments and tasks than the limited set of environments and activities that guided their evolution. To caricature, human bodies evolved for running in the East-African savanna, yet they are capable of climbing mount Everest, swimming across lakes, skydiving, playing basketball, etc. This is not a coincidence; by necessity, evolution optimizes for adaptability, whether cognitive adaptability or sensorimotor adaptability. Human physical capabilities can thus be said to be ``general'', but only in a limited sense; when taking a broader view, humans reveal themselves to be extremely specialized, which is to be expected given the process through which they evolved.

We argue that human cognition follows strictly the same pattern as human physical capabilities: both emerged as evolutionary solutions to specific problems in specific environments (commonly known as ``the four Fs''). Both were, importantly, optimized for adaptability, and as a result they turn out to be applicable for a surprisingly greater range of tasks and environments beyond those that guided their evolution (e.g. piano-playing, solving linear algebra problems, or swimming across the Channel) -- a remarkable fact that should be of the utmost interest to anyone interested in engineering broad or general-purpose abilities of any kind. Both are multi-dimensional concepts that can be modeled as a hierarchy of broad abilities leading up to a ``general'' factor at the top. And crucially, both are still ultimately highly specialized (which should be unsurprising given the context of their development): much like human bodies are unfit for the quasi-totality of the universe by volume, human intellect is not adapted for the large majority of conceivable tasks.

This includes obvious categories of problems such as those requiring long-term planning beyond a few years, or requiring large working memory (e.g. multiplying 10-digit numbers). This also includes problems for which our innate cognitive priors are unadapted; for instance, humans can be highly efficient in solving certain NP-hard problems of small size when these problems present cognitive overlap with evolutionarily familiar tasks such as navigation (e.g. the Euclidean Traveling Salesman Problem (TSP) with low point count can be solved by humans near-optimally in near-linear optimal time \cite{Macgregor1996}, using perceptual strategies), but perform poorly -- often no better than random search -- for problem instances of very large size or problems with less cognitive overlap with evolutionarily familiar tasks (e.g. certain non-Euclidean problems). For instance, in the TSP, human performance degrades severely when inverting the goal from ``finding the shortest path'' to ``finding the longest path'' \cite{Macgregor2011} -- humans perform even worse in this case than one of the simplest possible heuristic: farthest neighbor construction. \footnote{This does not necessarily mean that humanity as a collective is incapable of solving these problems; pooling individual humans over time or augmenting human intellect via external resources leads to increased generality, albeit this increase remains incremental, and still fundamentally differs from universality.}

A particularly marked human bias is dimensional bias: humans show excellent performance on 2D navigation tasks and 2D shape-packing puzzles, and can still handle 3D cases albeit with greatly reduced performance, but they are effectively unable to handle 4D and higher. This fact is perhaps unsurprising given human reliance on perceptual strategies for problem-solving -- strategies which are backed by neural mechanisms specifically evolved for 2D navigation (hippocampal systems of place cells and grid cells \cite{moser2015place}).

Thus, a central point of this document is that ``general intelligence'' is not a binary property which a system either possesses or lacks. It is a spectrum, tied to 1) a scope of application, which may be more or less broad, and 2) the degree of efficiency with which the system translate its priors and experience into new skills over the scope considered, 3) the degree of generalization difficulty represented by different points in the scope considered (see \ref{aFormalSynthesis}). In addition, the ``value'' of one scope of application over another is entirely subjective; we wouldn't be interested in (and wouldn't even perceive as intelligent) a system whose scope of application had no intersection with our own.

As such, it is conceptually unsound to set ``artificial general intelligence'' in an absolute sense (i.e. ``universal intelligence'') as a goal. To set out to build broad abilities of any kind, one must start from a target scope, and one must seek to achieve a well-defined intelligence threshold within this scope: AI is a deeply contextual and open-ended endeavour, not a single one-time riddle to be solved. However, it may in theory be possible to create human-like artificial intelligence: we may gradually build systems that extend across the same scope of applicability as human intelligence, and we may gradually increase their generalization power within this scope until it matches that of humans. We may even build systems with higher generalization power (as there is no \textit{a priori} reason to assume human cognitive efficiency is an upper bound), or systems with a broader scope of application. Such systems would feature intelligence beyond that of humans. 
\\

In conclusion, we propose that research on developing broad in AI systems (up to ``general'' AI, i.e. AI with a degree of generality comparable to human intelligence) should focus on \textit{defining, measuring, and developing a specifically human-like form of intelligence, and should benchmark progress specifically against human intelligence} (which is itself highly specialized). This isn't because we believe that intelligence that greatly differs from our own couldn't exist or wouldn't have value; rather, we recognize that \textit{characterizing and measuring intelligence is a process that must be tied to a well-defined scope of application, and at this time, the space of human-relevant tasks is the only scope that we can meaningfully approach and assess}. We thus disagree with the perspective of Universal Psychometrics \cite{HernandezOrallo2014} or Legg and Hutter's Universal Intelligence \cite{legg2007universal}, which reject anthropocentrism altogether and seek to measure all intelligence against a single absolute scale. An anthropocentric frame of reference is not only legitimate, it is necessary.

\subsubsection{Separating the innate from the acquired: insights from developmental psychology}
\label{separatingTheInnateFromTheAcquired}

Advances in developmental psychology teach us that neither of the two opposing views of the nature of the mind described in \ref{definingIntelligence} are accurate (see e.g. \cite{Spelke2007}): the human mind is not merely a collection of special-purpose programs hard-coded by evolution; it is capable of a remarkable degree of generality and open-endedness, going far beyond the scope of environments and tasks that guided its evolution. The large majority of the skills and knowledge we possess are acquired during our lifetimes, rather than innate. Simultaneously, the mind is not a single, general-purpose ``blank slate'' system capable of learning anything from experience. Our cognition is specialized, shaped by evolution in specific ways; we are born with \textit{priors} about ourselves, about the world, and about how to learn, which determine what categories of skills we can acquire and what categories of problems we can solve.

These priors are not a limitation to our generalization capabilities; to the contrary, they are their source, the reason why humans are capable of acquiring certain categories of skills with remarkable efficiency. The central message of the No Free Lunch theorem \cite{Wolpert1997} is that to learn from data, one must make assumptions about it -- the nature and structure of the innate assumptions made by the human mind are precisely what confers to it its powerful learning abilities.

We noted in \ref{measuringTheRightThing} that an actionable measure of intelligence should, crucially, control for \textit{priors} and \textit{experience}. We proposed in \ref{theMeaningOfGenerality} that evaluating general intelligence should leverage human intelligence as a necessary frame of reference. It follows that we need a clear understanding of \textit{human cognitive priors} in order to fairly evaluate general intelligence between humans and machines.

Human cognitive priors come in multiple forms, in particular \footnote{The boundaries between these categories may be fluid; the distinction between low-level sensorimotor priors and high-level knowledge priors is more one of degree than one of nature; likewise, the distinction between meta-learning priors and knowledge priors is subjective since knowledge facilitates skill acquisition: for instance, the neural mechanism behind our capabilities to perform 2D navigation may be treated either as a specialized meta-learning prior or as a knowledge prior about the external world.}:

\begin{itemize}
    \item Low-level priors about the structure of our own sensorimotor space, e.g. reflexes such as the vestibulo-ocular reflex, the palmar grasp reflex, etc. These priors enable infants (including prior to birth) to quickly take control of their senses and bodies, and may even generate simple behaviors in a limited range of situations.
    \item Meta-learning priors governing our learning strategies and capabilities for knowledge acquisition. This may include, for instance, the assumption that information in the universe follows a modular-hierarchical structure, as well as assumptions regarding causality and spatio-temporal continuity.
    \item High-level \textit{knowledge priors} regarding objects and phenomena in our external environment. This may include prior knowledge of visual objectness (what defines an object), priors about orientation and navigation in 2D and 3D Euclidean spaces, goal-directedness (expectation that our environment includes agents that behave according to goals), innate notions about natural numbers, innate social intuition (e.g. theory of mind), etc.
\end{itemize}

When it comes to creating artificial human-like intelligence, low-level sensorimotor priors are too specific to be of interest (unless one seeks to build an artificial human body). While human meta-learning priors should be of the utmost interest (understanding the strategies that the brain follows to turn experience into knowledge and skills is effectively our end goal), these priors are not relevant to evaluating intelligence: they \textit{are} intelligence, rather than a third-party modulating factor to be controlled for. They are part of the black box that we seek to characterize.

It is knowledge priors that should be accounted for when measuring a human-like form of intelligence. A system that does not possess human innate knowledge priors would be at a critical disadvantage compared to humans when it comes to efficiently turning a given experience curriculum into skill at a given human task. Inversely, a system that has access to more extensive hard-coded knowledge about the task at hand could not be fairly compared to human intelligence -- as we noted in \ref{measuringTheRightThing}, unlimited priors allow system developers to ``buy'' unbounded performance on any given task, with no implications with regard to generalization abilities (what we are actually trying to achieve).\\

Therefore, \textit{we propose that an actionable test of human-like general intelligence should be founded on innate human knowledge priors}:

\begin{itemize}
    \item The priors should be made as close as possible to innate human knowledge priors as we understand them. As our understanding of human knowledge priors improves over time, so should the test evolve.

    \item The test should assume that the system being measured possesses a specific set of priors. AI systems with more extensive priors should not be benchmarked using such a test. AI systems with fewer priors should be understood to be at a disadvantage.

    \item The priors assumed by the test should be explicitly and exhaustively described. Importantly, current psychometric intelligence tests make many assumptions about prior knowledge held by the test-taker (either innate or acquired), but never explicitly describe these assumptions.

    \item To make sure that humans test-takers do not bring further priors to the test, the test tasks should not rely on any acquired human knowledge (i.e. any knowledge beyond innate prior knowledge). For instance, they should not rely on language or learned symbols (e.g. arrows), on acquired concepts such as ``cat'' or ``dog'', or on tasks for which humans may have trained before (e.g. chess).

\end{itemize}

This leads us to a central question: what is the exact list of knowledge priors that humans are born with? This is the question that the developmental science theory of Core Knowledge \cite{Spelke2007} seeks to answer. Core Knowledge identifies four broad categories of innate assumptions that form the foundations of human cognition, and which are largely shared by our non-human relatives \footnote{Core Knowledge has been written into our DNA by natural evolution. Natural evolution is an extremely low-bandwidth, highly selective mechanism for transferring information from the surrounding environment to an organism's genetic code. It can only transfer information associated with evolutionary pressures, and it can only write about aspects of the environment that are stable over sufficiently long timescales. As such, it would not be reasonable to expect humans to possess vast amounts of human-specific prior knowledge; core knowledge is evolutionarily ancient and largely shared across many species, in particular non-human primates.}:

\begin{itemize}
    \item Objectness and elementary physics: humans assume that their environment should be parsed into ``objects'' characterized by principles of cohesion (objects move as continuous, connected, bounded wholes), persistence (objects do not suddenly cease to exist and do not suddenly materialize), and contact (objects do not act at a distance and cannot interpenetrate).
    \item Agentness and goal-directedness: humans assume that, while some objects in their environment are inanimate, some other objects are ``agents'', possessing intentions of their own, acting so as to achieve goals (e.g. if we witness an object A following another moving object B, we may infer that A is pursuing B and that B is fleeing A), and showing efficiency in their goal-directed actions. We expect that these agents may act contingently and reciprocally.
    \item Natural numbers and elementary arithmetic: humans possess innate, abstract number representations for small numbers, which can be applied to entities observed through any sensory modality. These number representations may be added or subtracted, and may be compared to each other, or sorted.
    \item Elementary geometry and topology: this core knowledge system captures notions of distance, orientation, in/out relationships for objects in our environment and for ourselves. It underlies humans' innate facility for orienting themselves with respect to their surroundings and navigating 2D and 3D environments.

\end{itemize}

While cognitive developmental psychology has not yet determined with a high degree of certainty the exact set of innate priors that humans possess, we consider the Core Knowledge theory to offer a credible foundation suitable to the needs of a test of human-like general intelligence. We therefore propose that an actionable test of general intelligence that would be fair for both humans and machines should only feature tasks that assume the four core knowledge systems listed above, and should not involve any acquired knowledge outside of these priors. We also argue, in agreement with \cite{LakeUTG16}, that general AI systems should hard-code as fundamental priors these core knowledge principles.

\subsection{Defining intelligence: a formal synthesis}
\label{aFormalSynthesis}

\subsubsection{Intelligence as skill-acquisition efficiency}
\label{intelligenceAsSkillAcquisitionEfficiency}

So far, we have introduced the following informally-described intuitions:

\begin{itemize}
    \item Intelligence lies in broad or general-purpose abilities; it is marked by flexibility and adaptability (i.e. skill-acquisition and generalization), rather than skill itself. The history of AI has been a slow climb along the spectrum of generalization.
    \item A measure of intelligence should imperatively control for experience and priors, and should seek to quantify generalization strength, since unlimited priors or experience can produce systems with little-to-no generalization power (or intelligence) that exhibit high skill at any number of tasks.
    \item Intelligence and its measure are inherently tied to a scope of application. As such, general AI should be benchmarked against human intelligence and should be founded on a similar set of knowledge priors.

\end{itemize}

Let us now formalize these intuitions. In what follows, we provide a series of definitions for key concepts necessary to ground a formal definition of intelligence and its measure. We will leverage the tools of Algorithmic Information Theory. These definitions lead up to a formal way of expressing the following central idea:\\

\textit{The intelligence of a system is a measure of its skill-acquisition efficiency over a scope of tasks, with respect to priors, experience, and generalization difficulty.}\\

Intuitively, if you consider two systems that start from a similar set of knowledge priors, and that go through a similar amount of experience (e.g. practice time) with respect to a set of tasks not known in advance, the system with higher intelligence is the one that ends up with greater skills (i.e. the one that has turned its priors and experience into skill more efficiently). This definition of intelligence encompasses meta-learning priors, memory, and fluid intelligence. It is distinct from skill itself: skill is merely the output of the process of intelligence.

Before we start, let us emphasize that many possible definitions of intelligence may be valid, across many different contexts, and we do not purport that the definition above and the formalism below represent the ``one true'' definition. Nor is our definition meant to achieve broad consensus. Rather, the purpose of our definition is to be actionable, to serve as a useful perspective shift for research on broad cognitive abilities, and to function as a quantitative foundation for new general intelligence benchmarks, such as the one we propose in part \ref{theARCDataset}. As per George Box's aphorism, \textit{``all models are wrong, but some are useful''}: our only aim here is to provide a useful North Star towards flexible and general AI. We discuss in \ref{practicalImplications} the concrete ways in which our formalism is useful and actionable.

\subsubsection*{Position of the problem}

First, we must introduce basic definitions to establish our problem setup. It should be immediately clear to the reader that our choice of problem setup is sufficient to model Fully-Supervised Learning, Partially-Supervised Learning, and Reinforcement Learning.

We consider the interaction between a ``task'' and an ``intelligent system''. This interaction is mediated by a ``skill program'' (generated by the intelligent system) and a  ``scoring function'' (part of the task).

We implicitly consider the existence of a fixed universal Turing machine on which our programs run (including the skill programs, as well as programs part of the task and part of the intelligent system). We also assume the existence of a fixed ``situation space'' $SituationSpace$ and ``response space'' $ResponseSpace$. Each of these spaces defines the set of binary strings that are allowed as input (and output, respectively) of all skill programs we will consider henceforth. They may be, for instance, the sensor space and the motor space of an animal or robot.
\begin{figure}[h]
    \centering
    \includegraphics[scale=0.36]{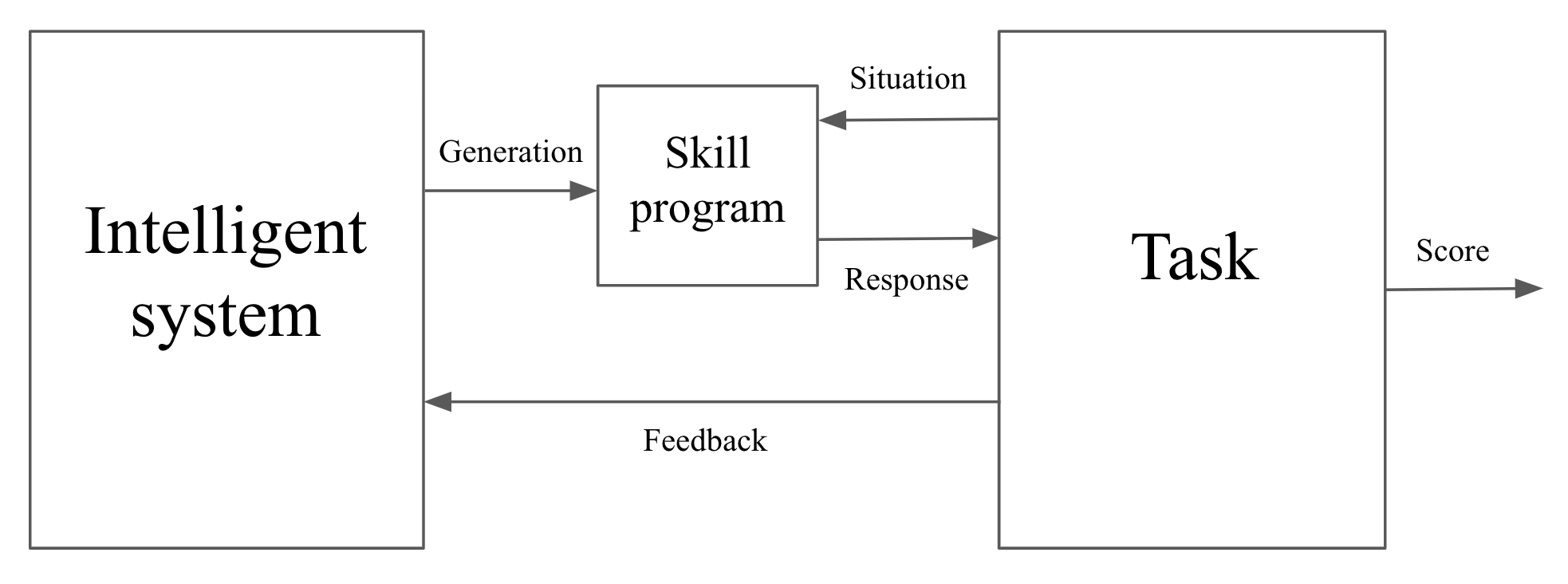}
    \caption{Position of the problem: an intelligent system generates a skill program to interact with a task.}
    \label{fig:position-of-the-problem}
\end{figure}
\\\\
A \textbf{task} $T$ consists of four objects:

\begin{itemize}
    \item A task state $TaskState$ (binary string).
    \item A ``situation generation'' function $SituationGen:TaskState \rightarrow Situation$. It may be stochastic.
        \begin{itemize}
            \item A $Situation$ is a binary string belonging to $SituationSpace$.
        \end{itemize} 
    \item A ``scoring function'' $Scoring:[Situation, Response, TaskState] \rightarrow [Score, Feedback]$. It may be stochastic.
        \begin{itemize}
            \item A $Response$ is a binary string belonging to $ResponseSpace$.
            \item A ``score'' $Score$ is a scalar. It is meant to measure the appropriateness of a response to a situation.
            \item A piece of ``feedback'' $Feedback$ is a binary string. It may encode full or partial information about the current score, or about scores corresponding to past responses (which may be known to the task state).
            \item Note: The parameter $Situation$ is technically optional since it may be known to the task state at runtime -- we include it here for maximum explicitness.
        \end{itemize}
    \item A self-update function $TaskUpdate: [Response, TaskState] \rightarrow TaskState$, which mutates the task state based on the response to the latest situation. It may be stochastic.
    
\end{itemize}

\noindent For instance, a game such as chess or WarCraft III (as well as what we call ``task'' in the ARC benchmark presented in \ref{theARCDataset}) would constitute a task. A given chess board position, screen frame in WarCraft III, or input grid in ARC, would constitute a situation.\\

\noindent An \textbf{intelligent system} $IS$ consists of three objects:

\begin{itemize}
    \item A system state $ISState$ (binary string).
    \item A ``skill program generation function'':\\
        $SkillProgramGen:ISState \rightarrow [SkillProgram, SPState]$.\\
        It may be stochastic.
        \begin{itemize}
            \item A skill program $SkillProgram:[Situation, SPState] \rightarrow [Response, SPState]$\\
            is a function that maps an input situation to a valid response (part of $ResponseSpace$), potentially using some working memory ($SPState$). It may be stochastic. Because it possesses a state $SPState$ (a binary string), it may be used to autonomously handle a series of connected situations without further communication with the intelligent system that generated it.
            \item A skill program may be, for instance, any game-specific program capable of playing new levels in a given video game.
            \item In what follows, we refer to ``skill program'' as the combination of the $SkillProgram$ function and the initial skill program state $SPState$ (i.e. skill programs are considered stateful).
            \item A skill program represents a frozen version of the system's task-specific capabilities (including the ability to adapt to novel situations within the task). We use the concept of skill program as a conceptual device to formalize the level of task-specific skill and task-specific generalization capabilities of an agent at a given point in time.
        \end{itemize}
    \item A self-update function $ISUpdate: [Situation, Response, Feedback, ISState] \rightarrow ISState$, which mutates the system's state based on the latest situation and corresponding feedback. It may be stochastic.
\end{itemize}

\noindent For instance, a neural network generation and training algorithm for games would be an ``intelligent system'', and the inference-mode game-specific network it would output at the end of a training run on one game would be a ``skill program''. A program synthesis engine capable of looking at an ARC task and outputting a solution program would be an ``intelligent system'', and the resulting solution program capable of handling future input grids for this task would be a ``skill program''.\\

\noindent The interaction between task, intelligent system, and skill programs is structured in two phases: a training phase and an evaluation phase. The goal of the training phase is for the $IS$ to generate a high-skill skill program that will generalize to future evaluation situations. The goal of the evaluation phase is to assess the capability of this skill program to handle new situations.\\

\noindent The training phase consists of the repetition of the following steps (we note the current step as $t$). Before we start, we consider two separate initial task states, $trainTaskState_{t=0}$ and $testTaskState_{e=0}$.

\begin{itemize}
    \item We generate a training situation: $situation_t \leftarrow SituationGen(trainTaskState_t)$
    \item The $IS$ generates a new skill program (without knowledge of the current situation):\\
        $[skillProgram_t, spState_t] \leftarrow SkillProgramGen(isState_t)$
        \begin{itemize}
            \item Implicitly, we assume that the ``goal'' of the $IS$ is to generate highly-skilled programs, i.e. programs that would have performed well on past situations, that will perform well on the next situation, and that would perform well on any possible situation for this task (in particular evaluation situations, which may feature significant novelty and uncertainty). We do not attempt to model \textit{why} the $IS$ should pursue this goal.
            \item $spState_t$ represents the working memory of the skill program at time $t$. Note that, because the skill program is generated anew with each training step, statefulness across situations via $SPState$ is not actually required during training. However, statefulness is important during evaluation when handling tasks that require maintaining information across situations. Note that in many games or real-world tasks, situations are all independent and thus skill programs don't require statefulness at all (e.g. ARC, or any fully-observable game, like chess).
        \end{itemize}
    \item The skill program outputs a response to the situation:\\
        $[response_t, spState_{t+1}] \leftarrow skillProgram_t(Situation_t, spState_t)$
        \begin{itemize}
            \item $skillProgram_t$ is only called once, and $spState_{t+1}$ is discarded, since the skill program is generated anew by the intelligent system at each training step.
            \item In practice, in partially-observable games where consecutive situations are very close to each other (e.g. two consecutive screen frames in WarCraft III), one may assume that $skillProgram_t$ at $t$ and $skillProgram_{t+1}$ would not actually be generated independently from scratch and would stay very close to each other (i.e. the $IS$'s understanding of the task would be evolving continuously in program space); $spState_{t+1}$ as generated by $skillProgram_t$ and $spState_{t+1}$ as generated by $SkillProgramGen$ at $t+1$ would likewise stay very close to each other.
        \end{itemize}
    \item The task scoring function assigns a score to the response and generates a piece of feedback:
        $[score_t, feedback_t] \leftarrow Scoring(Situation_t, response_t, trainTaskState_t)$
        \begin{itemize}
            \item Note: The scalar score is meant to encode how appropriate the response is, and the feedback data is meant to be used by the intelligent system to update its state. In simple cases (e.g. fully-supervised learning), the feedback data is the same as the scalar score, meaning that the intelligent agent would have complete and immediate information about the appropriateness of its response. In other cases, the feedback data may only contain partial information, no information, or information that is only relevant to responses generated for prior situations (delayed feedback).
        \end{itemize}
    \item The $IS$ updates its internal state based on the feedback received from the task:\\
        $isState_{t+1} \leftarrow ISUpdate(Situation_t, response_t, feedback_t, isState_t)$
    \item The task updates its internal state based on the response received to the situation:\\
        $trainTaskState_{t+1} \leftarrow TaskUpdate(response_t, trainTaskState_t)$
\end{itemize}

\noindent The training phase ends at the discretion of the $SituationGen$ function (e.g. $SituationGen$ returns a ``STOP'' situation), at which time $SkillProgramGen$ would generate its last skill program, including an initial state (initial working memory) meant to perform well during evaluation (e.g. blank).\\

\noindent The evaluation phase is superficially similar to the training phase, with the differences that 1) the task starts from $testTaskState_{e=0}$ and consists of an independent series of situations, 2) it only involves a single fixed skill program $testSkillProgram$ starting with state $testSPState_{e=0}$. Crucially, it no longer involves the intelligent system. Note that $testTaskState_{e=0}$ could be chosen stochastically. For instance, different randomly chosen initial $testTaskState_{e=0}$ could be different randomly-generated levels of a game.\\

\noindent Like the separation between skill program and intelligent system, the evaluation phase should be understood as a conceptual device used to quantify the task-specific skill and task-specific generalization capabilities demonstrated by a system at a given point in time. The evaluation phase should \textit{not} be seen as being conceptually similar to a child taking a school test or an IQ test. In real-world evaluation situations, evaluation involves the entire intelligent system, dynamically adapting its understanding of the task at hand. A real-world evaluation situation would be represented in our formalism as being part of the training curriculum -- a series of training situations with blank feedback.\\

\noindent The evaluation phase consists of the repetition of the following steps (the current step is noted $e$):

\begin{itemize}
    \item We generate a test situation: $situation_e \leftarrow SituationGen(testTaskState_e)$
    \item The skill program considered produces a response:\\
        $[response_e, testSPState_{e+1}] \leftarrow testSkillProgram(situation_e, testSPState_e)$
        \begin{itemize}
            \item Note that the update of the skill program state enables the skill program to maintain a working memory throughout the evaluation phase. This is useful for partially-observable games. This is irrelevant to many games (including ARC) and many real-world tasks, where skill programs would be stateless.
        \end{itemize}
    \item The task scoring function assigns a score to the response (the feedback is discarded):
        $score_e \leftarrow Scoring(Situation_e, response_e, testTaskState_e)$
    \item The task updates its internal state based on the response received:\\
        $testTaskState_{e+1} \leftarrow TaskUpdate(response_e, testTaskState_e)$
\end{itemize}

\noindent The evaluation phase also ends at the discretion of the $SituationGen$ function.\\

\noindent Note that for the sake of simplification, we consider that the $IS$'s state does not transfer across tasks; the $IS$ would start with a ``blank'' state at the beginning of the training phase for each new task (i.e. only possessing built-in priors). However, the setup above and definitions below may be readily extended to consider lifelong learning, to bring it closer to real-world biological intelligent systems, which learn continuously across a multiplicity of partially overlapping tasks with often no clear boundaries.\\

\noindent Based on the setup described thus far, we can define the following useful concepts:

\begin{itemize}
    \item \textit{Evaluation result}: Sum of the scalar scores obtained by a fixed skill program over a specific evaluation phase instance for a task. Since all objects involved (skill program, situation generation program, task update program, initial task state) may be stochastic, this quantity may also be stochastic. Likewise, we define \textit{training-time performance} as the sum of the scalar scores obtained during a given training phase. Training-time performance is tied to a specific sequence of training situations.

    \item \textit{Skill}: Probabilistic average of evaluation results over all possible evaluation phase instances, i.e. average of per-evaluation sum of scores obtained after running the evaluation phase infinitely many times. Skill is a property of a skill program. Note that other distributional reduction functions could be used, such as median or minimum.
    
    \item \textit{Optimal skill}: Maximum skill theoretically achievable by the best possible skill program on the task. It is a property of a task.
    
    \item \textit{Sufficient skill threshold}, noted $\theta_{T}$: Subjective threshold of skill associated with a task, above which a skill program can be said to ``solve'' the task. It is a property of a task.
    
    \item \textit{Task and skill value function}: We define a value function over task space (note that task space may be infinite), associating a scalar value to the combination of a task and a threshold of skill $\theta$ for the task : $TaskValue: Task, \theta \rightarrow \omega_{T, \theta}$. Values are assumed positive or zero, and $TaskValue$ is assumed monotonous as a function of $\theta$ (for a given task, higher skill always has higher value). This value function captures the relative importance of skill at each task and defines the subjective frame of reference of our intelligence definition (for instance, if we wish to evaluate human-like intelligence, we would place high value on achieving high skill at human-relevant tasks and place no value on tasks that are irrelevant to the human experience). The value $\omega_{T, \theta}$ of a skill level at a task is chosen so that the quantity $\omega_{T, \theta}$ can be compared fairly across different tasks (i.e. it should capture the value we place on achieving skill $\theta$ at task $T$). This enables us to homogeneously aggregate skill across different tasks without worrying about the scale of the their respective scoring functions.
    
    \item \textit{Task value}, noted $\omega_{T}$: This is the value of achieving sufficient skill level at $T$, i.e. $\omega_{T} = \omega_{T, \theta_{T}}$.
    
    \item \textit{Optimal solution}: Any skill program that can achieve optimal skill on a task. Likewise we define a \textit{training-time optimal solution} as any skill program that can achieve optimal training-time performance over a specific sequence of training situations.
    
    \item \textit{Sufficient solution}: Any skill program that can achieve sufficient skill $\theta_{T}$ on a task.
    
    \item \textit{Curriculum}: Sequence of interactions (situations, responses, and feedback) between a task and an intelligent system over a training phase. For a given task and intelligent system, there exists a space of curricula, parameterized by the stochastic components of the underlying programs. A curriculum emerges from the interaction between the system and a task: this can model both teaching and active learning.

    \item \textit{Optimal curriculum}: curriculum which leads an intelligent system to produce the best (highest skill) skill program it can generate for this task. It is specific to a task and an intelligent system. There may be more than one optimal curriculum.

    \item \textit{Sufficient curriculum}: curriculum which leads an intelligent system to a sufficient solution. It is specific to a task and an intelligent system. There may be more than one sufficient curriculum.

    \item \textit{Task-specific potential}, noted $\theta^{max}_{T, IS}$: Skill of the best possible skill program that can be generated by a given intelligent system on a task (after an optimal curriculum). It is a scalar value specific to a task and an intelligent system.
    
    \item \textit{Intelligent system scope}: Subspace of task space including all tasks for which task value $\omega_{T}$ is non-zero and for which the intelligent system is capable of producing a sufficient solution after a training phase. This space may be infinite. ``To be capable of producing a sufficient solution'' means that there exists a sufficient curriculum for the intelligent system and task considered. A scope is a property of an intelligent system.

    \item \textit{Intelligent system potential}: Set of task-specific potential values over all tasks in the system's scope. Potential is a property of an intelligent system.

\end{itemize}

\noindent We find that in most cases it is more useful to consider sufficient skill and sufficient solutions rather than optimal skill and optimal solutions -- in application settings, we seek to achieve sufficient performance using as little resources as possible; it is rarer and less practical to seek to achieve maximum possible performance using unlimited resources.

\subsubsection*{Quantifying generalization difficulty, experience, and priors using Algorithmic Information Theory}

Algorithmic Information Theory (AIT) may be seen as a computer science extension of Information Theory. AIT concerns itself with formalizing useful computer science intuitions regarding complexity, randomness, information, and computation. Central to AIT is the notion of Algorithmic Complexity. Algorithmic Complexity (also known as Kolmogorov Complexity or Algorithmic Entropy) was independently investigated, in different contexts, by R.J. Solomonoff, A.N. Kolmogorov and G.J. Chaitin in the 1960s. For an extensive introduction, see \cite{chaitin1975theory, Chaitin1987, GruenwaldVitanyi2008, li2008introduction}.

Much like the concept of Entropy in Information Theory, Algorithmic Complexity is a measure of the ``information content'' of mathematical objects. For our own needs, we will only consider the specific case of binary strings. Indeed, all objects we have introduced so far have been either scalar values (score, potential), or binary strings (states, programs, situations, and responses), since any program may be represented as a binary string.

The Algorithmic Complexity (noted $H(s)$) of a string $s$ is the length of the shortest description of the string in a fixed universal language, i.e. the length of the shortest program that outputs the string when running on a fixed universal Turing machine. Since any universal Turing machine can emulate any other universal Turing machine, $H(s)$ is machine-independent to a constant.

We can use Algorithmic Complexity to define the information content that a string $s_2$ possesses about a string $s_1$ (called ``Relative Algorithmic Complexity'' and noted $H(s_1|s_2)$), as the length of the shortest program that, taking $s_2$ as input, produces $s_1$. ``To take $s_2$ as input'' means that $s_2$ is part of the description of the program, but the length of $s_2$ would not be taken into account when counting the program's length.

Because any program may be represented as a binary string, we can use Relative Algorithmic Complexity to describe how closely related two programs are. Based on this observation, we propose to define the intuitive notion of ``Generalization Difficulty" of a task as follows:\\

\noindent Consider:

\begin{itemize}
    \item A task $T$,
    \item $Sol^{\theta}_T$, the shortest of all possible solutions of $T$ of threshold $\theta$ (shortest skill program that achieves at least skill $\theta$ during evaluation),
    \item $TrainSol^{opt}_{T, C}$, the shortest optimal training-time solution given a curriculum (shortest skill program that achieves optimal training-time performance over the situations in the curriculum).
\end{itemize}

\noindent We then define Generalization Difficulty as:\\

\noindent \textbf{Generalization Difficulty of a task given a curriculum $C$ and a skill threshold $\theta$}, noted $GD^{\theta}_{T, C}$:
Fraction of the Algorithmic Complexity of solution $Sol^{\theta}_T$ that is explained by the shortest optimal training-time solution $TrainSol^{opt}_{T, C}$ (i.e. length of the shortest program that, taking as input the shortest possible program that performs optimally over the situations in curriculum $C$, produces a program that performs at a skill level of at least $\theta$ during evaluation, normalized by the length of that skill program). Note that this quantity is between 0 and 1 by construction.\\
\item $GD^{\theta}_{T, C} = \frac{H(Sol^{\theta}_T|TrainSol^{opt}_{T, C})}{H(Sol^{\theta}_T)}$\\

\noindent Thus, a task with high ``generalization difficulty'' is one where the evaluation-time behavior needs to differ significantly from the simplest possible optimal training-time behavior in order to achieve sufficient skill. Relative Algorithmic Complexity provides us with a metric to quantify this difference: $GD$ is a measure of how much the shortest training-time solution program needs to be edited in order to become an appropriate evaluation-time solution program. If the shortest skill program that performs optimally during training also happens to perform at a sufficient skill level during evaluation, the task has zero generalization difficulty (i.e. it does not involve uncertainty). A generalizable skill program is one that ``covers more ground'' in situation space than the exact training situations it is familiar with: a program that is capable of dealing with future uncertainty.

Note that this definition of generalization difficulty may seem counter-intuitive. Occam's razor principle would seem to suggest that the simplest program that works on the training situations should also be a program that generalizes well. However, generalization describes the capability to deal with future uncertainty, not the capability to compress the behavior that would have been optimal in the past -- being prepared for future uncertainty has a cost, which is antagonistic to policy compression \footnote{As a philosophical aside: this is why the education of children involves practicing games and ingesting knowledge of seemingly no relevance to their past or present decision-making needs, but which prepare them for future situations (a process often driven by curiosity). A 10-year-old who has only learned the simplest behavioral policy that would have maximized their extrinsic rewards (e.g. candy intake) during ages 0-10 would not be well educated, and would not generalize well in future situations.}. By necessity, $TrainSol^{opt}_{T, C}$ does away with any information or capability that isn't strictly necessary in order to produce the correct response to training situations, and in doing so, it may discard information or capabilities that would have been useful to process evaluation situations. If it is in fact the case that $TrainSol^{opt}_{T, C}$ does not need to discard any such information (i.e. the simplest behavior that was optimal in the past is still sufficient in the future), this implies that the evaluation features no need for adaptation (no  non-trivial novelty or uncertainty), and thus the task does not involve generalization, potentially given some starting point (such as the solution of another task).

Another way to express the same idea is that \textit{generalization requires to reinterpret the task when new data arrives} (e.g. at evaluation time). This implies the need to store representations of past data that would be seemingly useless from the perspective of the past but may prove useful in the future. For example, consider the following labeled points along a line: $(x=-0.75, label=False), (x=0.15, label=True), (x=-0.1, label=True)$. When training a classification program on the first two of these points, some of the shortest optimal training-time solutions may be $\lambda(x): x > 0$ or $\lambda(x): bool(ceil(x))$. When applied to the last point $(x=-0.1, label=True)$, these solutions would fail, while an algorithm that instead stores all past data points and uses nearest-neighbors to return a response at evaluation time would work. The nearest-neighbors program would be better prepared for future uncertainty, but would take significantly more space to write down.\\

\noindent Importantly, this first definition of generalization difficulty only captures system-centric generalization, as it quantifies the difficulty of handling evaluation situations that differ from training situations regardless of the system's preexisting capabilities. To capture developer-aware generalization, we need to take into account the system in its initial state at the start of training, ${SkillProgramGen, ISUpdate, isState_{t=0}}$:\\

\noindent\textbf{Developer-aware Generalization Difficulty of a task for an intelligent system given a curriculum $C$ and a skill threshold $\theta$}, noted $GD^{\theta}_{IS, T, C}$: Fraction of the Algorithmic Complexity of solution $Sol^{\theta}_T$ that is explained by $TrainSol^{opt}_{T, C}$ and the initial state of the system $IS_{t=0}$, i.e. length of the shortest program that, taking as input the initial system plus the shortest possible program that performs optimally over the situations in curriculum $C$, produces a skill program that performs at a skill level of at least $\theta$ during evaluation, normalized by the length of that skill program. Note that this quantity is between 0 and 1 by construction.\\
\item $GD^{\theta}_{IS, T, C} = \frac{H(Sol^{\theta}_T|TrainSol^{opt}_{T, C}, IS_{t=0})}{H(Sol^{\theta}_T)}$\\

\noindent In which we note: $IS_{t=0} = SkillProgramGen, ISUpdate, isState_{t=0}$\\

\noindent Developer-aware generalization thus represents the amount of uncertainty about the shortest evaluation-time solution given that you have at your disposal both the initial system and the shortest training-time solution, i.e. the amount of modifications you would have to make to the shortest training-time solution to obtain the evaluation-time solution, provided that these edits can make use of the contents of the initial system.\\

\noindent Likewise, we can define the Generalization Difficulty from task $T_1$ to task $T_2$ (sufficient case) as $H(Sol^{\theta_{T_2}}_{T_2}| Sol^{\theta_{T_1}}_{T_1}) / H(Sol^{\theta_{T_2}}_{T_2})$. We can also extend these definitions to a set of tasks (e.g. Generalization Difficulty from a set of practice task to a set of test tasks), which can be useful to quantify the Generalization Difficulty of an entire test suite. These notions are related to the concept of intrinsic task difficulty (regardless of generalization) defined in \cite{MeasureOfAllMinds2017} (section 8.6) as the effort necessary to construct a solution.\\

\noindent Next, we can also use Relative Algorithmic Complexity to formally quantify the \textbf{Priors} $P_{IS, T}$ possessed by an intelligent system about a task:\\

\noindent\textbf{Priors of an intelligent system relative to a task $T$ and a skill threshold $\theta$}, noted $P^{\theta}_{IS, T}$: Fraction of the Algorithmic Complexity of the shortest solution of $T$ of skill threshold $\theta$ that is explained by the initial system (at the start of the training phase). This is the length (normalized by $H(Sol^{\theta}_T)$) of the shortest possible program that, taking as input the initial system
${SkillProgramGen, ISUpdate, isState_{t=0}}$ (noted $IS_{t=0}$), produces the shortest  solution of $T$ that performs at a skill level of at least $\theta$ during evaluation. Note that the intelligent system does not need to be able to produce this specific solution. Note that this quantity is between 0 and 1 by construction.\\
\item $P^{\theta}_{IS, T} = \frac{H(Sol^{\theta}_T) - H(Sol^{\theta}_T|IS_{t=0})}{H(Sol^{\theta}_T)}$\\

\noindent ``Priors'' thus defined can be interpreted as a measure of how close from a sufficient or optimal solution the system starts, i.e. the ``amount of \textit{relevant} information'' embedded in the initial system. Note that this is different from the ``amount of information'' embedded in the initial system (which would merely be the Algorithmic Complexity of the initial system). As such, our measure only minimally penalizes large systems that contain prior knowledge that is irrelevant to the task at hand (the only added cost is due to knowledge indexing and retrieval overhead).\\

\noindent Further, we can use Relative Algorithmic Complexity to define the \textbf{Experience} $E_{IS, T, C}$ accumulated by an intelligent system about a task during a curriculum.\\

\noindent Consider a single step $t$ during training:

\begin{itemize}
    \item At $t$, the system receives some new \textit{data} in the form of the binary strings $situation_t$, $response_t$, and $feedback_t$ (although $response_t$ may be omitted since, being the output of a skill program previously generated by the $IS$, it can be assumed to be known by the $IS$ as soon as $situation_t$ is known).
    \item Only some of this data is \textit{relevant} to solving the task (the data may be noisy or otherwise uninformative).
    \item Only some of the data contains \textit{novel} information for the intelligent system (situations and responses may be repetitive, and the intelligent system may be a slow learner that needs information to be repeated multiple times or presented in multiple ways). Note that we use the term ``novel'' to characterize information that would appear novel to the system, rather than information that has never appeared before in the curriculum (the difference between the two lies in the system's learning efficiency).
\end{itemize}

\noindent We informally define the amount of experience accrued at step $t$ as the \textit{amount of relevant, novel information received by the system at $t$}. This corresponds to the amount of potential uncertainty reduction about the solution that is made available by the task in the current situation data and feedback data (i.e. how much the $IS$ could reduce its uncertainty about the solution using the step data if it were optimally intelligent).\\

\noindent Formally:\\

\noindent \textbf{Experience accrued at step $t$}, noted $E_{IS, T, t}^{\theta}$:\\

$E_{IS, T, t}^{\theta} = H(Sol^{\theta}_T | IS_t) - H(Sol^{\theta}_T | IS_t, data_t)$\\

\noindent In which we note:

\begin{itemize}
    \item $IS_t = SkillProgramGen, ISUpdate, isState_t$
    \item $data_t = Situation_t, response_t, feedback_t$
\end{itemize}

\noindent By summing over all steps, we obtain the following definition of total experience (note that we normalize by the Algorithmic Complexity of the solution considered, as we did for priors):\\

\noindent \textbf{Experience $E_{IS, T, C}^{\theta}$ over a curriculum $C$}:\\

$E_{IS, T, C}^{\theta} = \frac{1}{H(Sol^{\theta}_T)}\sum\limits_{t}{E_{IS, T, t}^{\theta}}$\\

\noindent ``Experience'' thus defined can be interpreted as a measure of the amount of \textit{relevant} information received by the system about the task over the course of a curriculum, only accounting for novel information at each step.

Because this is different from the ``amount of information'' contained in the curriculum (i.e. the Algorithmic Complexity of the curriculum), our measure does not penalize systems that go through noisy curricula.

In addition, because we use an eager sum of relevant and novel information at each step instead of globally pooling the information content of the curriculum, we penalize learners that are slower to absorb the relevant information that is presented to them.

Lastly, because our sum is different from ``amount of relevant information (novel or not) at each step summed over all steps'', we do not penalize systems that go through repetitive curricula. If a fast learner absorbs sufficient information over the first ten steps of a fixed curriculum, but a slow learner needs 90 more steps of the same curriculum to achieve the same, we will not count as experience for the fast learner the redundant last 90 steps during which it did not learn anything, but we will count all 100 steps for the slow learner.

\subsubsection*{Defining intelligence}

We have now established sufficient context and notations to formally express the intuitive definition of intelligence stated earlier, \textit{``the intelligence of a system is a measure of its skill-acquisition efficiency over a scope of tasks, with respect to priors, experience, and generalization difficulty.''}\\

\noindent We consider an intelligent system $IS$. We note $Cur_{T}^{\theta_{T}}$ the space of curricula that result in $IS$ generating a solution of sufficient skill $\theta_{T}$ for a task $T$, and $Cur_{T}^{opt}$ the space of curricula that result in $IS$ generating its highest-skill solution (solution reaching the system's potential $\theta^{max}_{T, IS}$). Note that system's potential may be lower than the optimal solution for the task, as the system may not be able to learn to optimally solve the task.

To simplify notations, we will denote $\theta^{max}_{T, IS}$ as $\Theta$. We note $Avg$ the averaging function (used to average over task space). We note $P_C$ the probability of a given curriculum $C$.\\

\noindent We then define the intelligence of $I$, tied to a scope of tasks $scope$, as:\\

\noindent \textbf{Intelligence of system $IS$ over $scope$ (sufficient case):}
    \begingroup
    \item \Large
        $I_{IS, scope}^{\theta_{T}} = \underset{T \in scope}{Avg} \left[ \omega_{T} \cdot \theta_{T} \underset{C \in Cur_{T}^{\theta_{T}}}{\Sigma} \left[ P_C \cdot \frac{ GD^{\theta_{T}}_{IS, T, C}}{P^{\theta_{T}}_{IS, T} + E_{IS, T, C}^{\theta_{T}}} \right] \right]$\\
    \endgroup
\\\\
\noindent \textbf{Intelligence of system $IS$ over $scope$ (optimal case):}
    \begingroup
    \item \Large
       $I_{IS, scope}^{opt} = \underset{T \in scope}{Avg} \left[ \omega_{T, \Theta} \cdot \Theta \underset{C \in Cur_{T}^{opt}}{\Sigma} \left[ P_C \cdot \frac{ GD^{\Theta}_{IS, T, C}}{P^{\Theta}_{IS, T} + E_{IS, T, C}^{\Theta}} \right] \right]$
    \endgroup
\\\\
\noindent Note that:
\begin{itemize}
    \item $P_{IS, T} + E_{IS, T, C}$ (priors plus experience) represents the total exposure of the system to information about the problem, including the information it starts with at the beginning of training.

    \item The sum over a curriculum subspace, weighted by the probability of each curriculum, represents the expected outcome for the system after training. Note that the sum is over a subspace of curricula (curricula that lead to at least a certain skill level), and thus the probabilities would sum to a total lower than one: as such, we are penalizing learners that only reach sufficient skill or optimal skill some of the time.
    
    \item $\omega_{T} \cdot \theta_{T}$ represents the subjective value we place on achieving sufficient skill at $T$, and $\omega_{T, \Theta} \cdot \Theta$ represents the subjective value we place on achieving the skill level corresponding to the system's full potential $\theta^{max}_{T, IS}$ at $T$.
    
    \item Schematically, the contribution of each task is: $Expectation \left[ \frac{skill \cdot generalization }{priors + experience} \right]$, which is further weighted by a value $\omega$ which enables us to homogeneously compare skill at different tasks independently of the scale of their respective scoring functions.
\end{itemize}

Thus, we equate the intelligence of a system to a measure of the information-efficiency with which the system acquires its final task-specific skill (sufficient skill or highest possible skill) on average (probabilistic average over all applicable curricula), weighted by the developer-aware generalization difficulty of the task considered (as well as the task value $\omega$, which makes skill commensurable across tasks), averaged over all tasks in the scope.

Or, in plain English: intelligence is the rate at which a learner turns its experience and priors into new skills at valuable tasks that involve uncertainty and adaptation.\\

\noindent Note that our definition is not the first formal definition of intelligence based on Algorithmic Information Theory. We are aware of three other AIT-based definitions: the C-Test \cite{HO98}, the AIXI model \cite{hutter2004universal}, and the ``Universal Intelligence'' model \cite{legg2007universal} (closely related to AIXI). It should be immediately clear to a reader familiar with these definitions that our own approach represents a very different perspective.\\

\begin{figure}[h]
    \centering
    \includegraphics[scale=0.36]{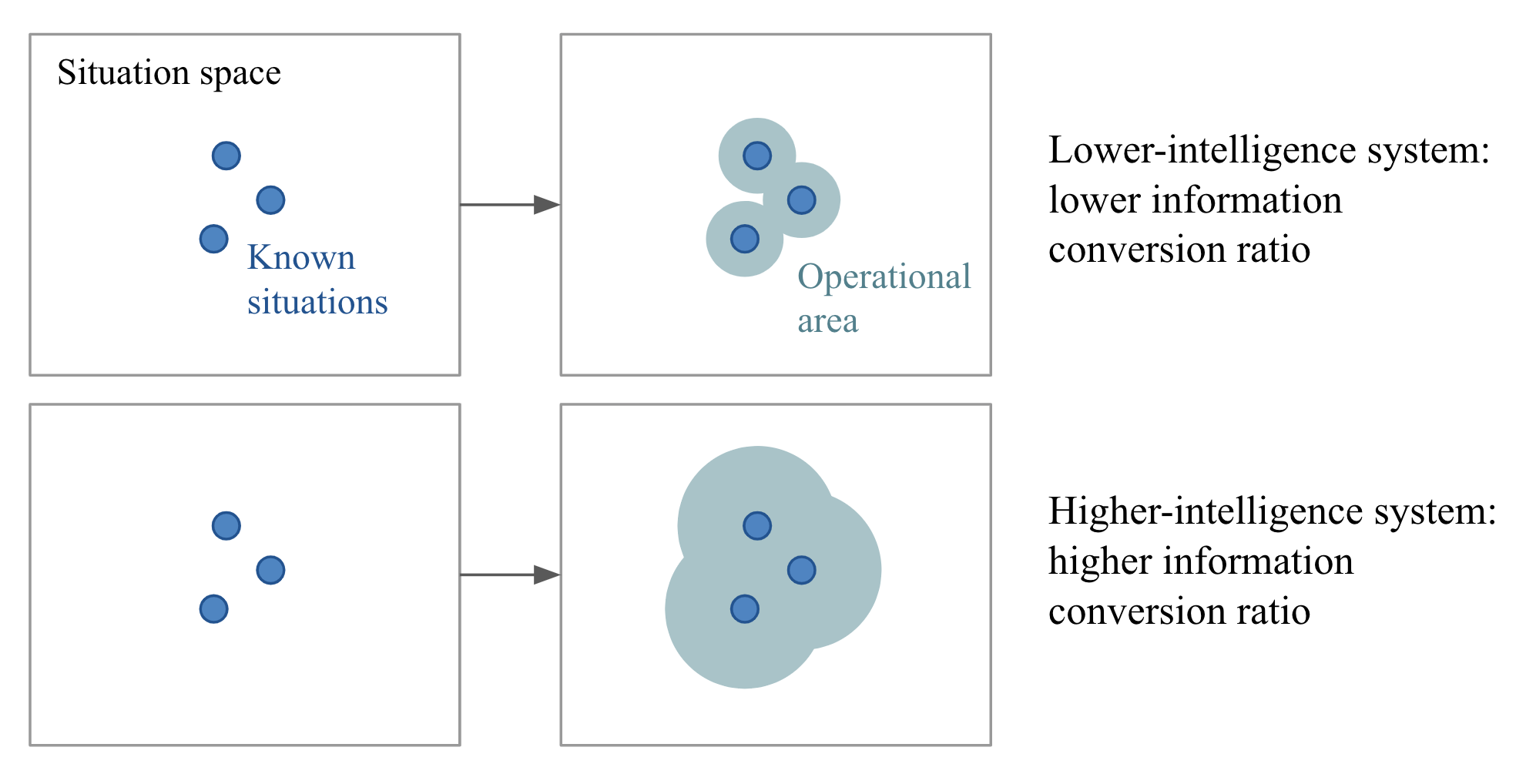}
    \caption{Higher intelligence ``covers more ground'' in future situation space using the same information.}
    \label{fig:conversion-ratio}
\end{figure}

\noindent We bring the reader's attention to a number of key observations about our formalism (see also \ref{practicalImplications}):

\begin{itemize}
    \item A high-intelligence system is one that can generate high-skill solution programs for high generalization difficulty tasks (i.e. tasks that feature high uncertainty about the future) using little experience and priors, i.e. it is a system capable of making highly efficient use of all of the information it has at its disposition to cover as much ground as possible in unknown parts of the situation space. Intelligence is, in a way, a conversion rate between information about part of the situation space, and the ability to perform well over a maximal area of future situation space, which will involve novelty and uncertainty (figure \ref{fig:conversion-ratio}).

    \item The measure of intelligence is tied to a choice of scope (space of tasks and value function over tasks). It can also optionally be tied to a choice of sufficient skill levels across the tasks in the scope (sufficient case).

    \item Skill is not possessed by an intelligent system, it is a property of the output artifact of the process of intelligence (a skill program). High skill is not high intelligence: these are different concepts altogether.
    
    \item Intelligence must involve learning and adaptation, i.e. operationalizing information extracted from experience in order to handle future uncertainty: a system that starts out with the ability to perform well on evaluation situations for a task would have a very low developer-aware generalization difficulty for this task, and thus would score poorly on our intelligence metric.
    
    \item Intelligence is not curve-fitting: a system that merely produces the simplest possible skill program consistent with known data points could only perform well on tasks that feature zero generalization difficulty, by our definition. An intelligent system must generate behavioral programs that account for future uncertainty.

    \item The measure of intelligence is tied to curriculum optimization: a better curriculum space will lead to greater realized skill (on average) and to greater expressed intelligence (greater skill-acquisition efficiency).
\end{itemize}

\subsubsection{Computation efficiency, time efficiency, energy efficiency, and risk efficiency}
\label{otherEfficiencies}

In the above, we only considered the information-efficiency (prior-efficiency and experience-efficiency with respect to generalization difficulty) of intelligent systems. Indeed, we believe this is the most actionable and relevant angle today to move AI research forward (cf \ref{practicalImplications}). But it isn't the only angle one may want to consider. Several alternatives that could be incorporated into our definition in various ways (e.g. as a regularization term) come to mind:

\begin{itemize}
    \item Computation efficiency of skill programs: for settings in which training data is abundant but inference-time computation is expensive, one may want to encourage the generation of the skill programs that have minimal computational resource consumption.
    \item Computation efficiency of the intelligent system: for settings in which training-time computation is expensive, one may want to expend a minimal amount of computation resources to generate a skill program.
    \item Time efficiency: in time-constrained settings, one may want to minimize the latency with which the intelligent system generates skill programs.
    \item Energy efficiency: in biological systems in particular, one may want to minimize the amount of energy expended in producing a skill program, in running a skill program, or in going through a curriculum.
    \item Risk efficiency: for settings in which going through a curriculum (i.e. collecting experience) involves risk for the intelligent system, one might want to encourage safe curricula at the expense of resource efficiency or information efficiency. Much like energy efficiency, this is highly relevant to biological systems and natural evolution, in which certain novelty-seeking behaviors that would lead to faster learning may also be more dangerous.

\end{itemize}

In fact, one may note that information efficiency acts in many settings as a proxy for energy efficiency and risk efficiency.

We expect that these alternative ways to quantify efficiency will become relevant in specialized AI application contexts in the future, and we bring them to the reader's attention to encourage others to develop new formal definitions of intelligence incorporating them in addition to information efficiency.

\subsubsection{Practical implications}
\label{practicalImplications}

The definitions above provide a formal framework as well as quantitative tools to reason about the intuitive notions we have been introducing so far, in particular the concepts of ``generalization difficulty'', ``intelligence as skill-acquisition efficiency'', and what it means to control for priors and experience when evaluating intelligence, as opposed to looking purely at task-specific skill.

The main value of this framework is to provide an actionable perspective shift in how we understand and evaluate flexible or general artificial intelligence. We argue that this perspective shift has the following practical consequences:

\paragraph*{a. Consequences for research directions towards flexible or general AI:}

\begin{itemize}
    \item It clearly spells out that the process of creating an intelligent system can be approached as an optimization problem, where the objective function would be a computable approximation of our quantitative intelligence formula. As pointed out in \ref{otherEfficiencies}, this objective function could be further refined by incorporating regularization terms that would take into account alternative forms of efficiency.
    \item It encourages a focus on developing broad or general-purpose abilities rather than pursuing skill alone, by proposing a target metric that penalises excessive reliance on experience or priors, and discounting tasks that feature low generalization difficulty.
    \item It encourages interest in program synthesis, by suggesting that we stop thinking of ``agents'' as monolithic black boxes that take in sensory input and produce behavior (a vision inherited from Reinforcement Learning \cite{RL}): our formalism clearly separates the part of the system that possesses intelligence (``intelligent system'', a program-synthesis engine) from the part that achieves skill or implements behavior (``skill program'', the non-intelligent output artifact of the process of intelligence), and places focus to the former. As we point out throughout this paper, we believe that this confusion between process and artifact has been an ongoing fundamental issue in the conceptualization of AI.
    \item It encourages interest in curriculum development, by leveraging the notion of an ``optimal curriculum'' and drawing attention to the fact that a better curriculum increases the intelligence manifested by a learning system.
    \item It encourages interest in building systems based on human-like knowledge priors (e.g. Core Knowledge) by drawing attention to the importance of priors in evaluating intelligence.

\end{itemize}

\paragraph*{b. Consequences for evaluating flexible or general AI systems:}

\begin{itemize}
    \item By defining and quantifying generalization difficulty, it offers a way to formally reason about what it means to perform ``local generalization'', ``broad generalization'', and ``extreme generalization'' (cf. the spectrum of generalization introduced in \ref{spectrumOfGeneralization}), and to weed out tests that feature zero generalization difficulty.
    \item It suggests concrete guidelines for comparing AI and human intelligence: such a comparison requires starting from a shared scope of tasks and shared priors, and would seek to compare experience-efficiency in achieving specific levels of skill. We detail this idea in \ref{fairComparisons}.
    \item It shows the importance of taking into account generalization difficulty when developing a test set to evaluate a task. We detail this idea in \ref{whatToExpectOfAnIdealBenchmark}. This should hopefully lead us to evaluation metrics that are able to discard solutions that rely on shortcuts that do not generalize (e.g. reliance on local textures as opposed to global semantics in computer vision).
    \item It provides a set of practical questions to ask about any intelligent system to rigorously characterize it:
        \begin{itemize}
            \item What is its scope?
            \item What is its ``potential'' over this scope (maximum achievable skill)?
            \item What priors does it possess?
            \item What is its skill-acquisition efficiency (intelligence)?
            \item What curricula would maximize its skill or skill-acquisition efficiency?
        \end{itemize}
\end{itemize}

\subsection{Evaluating intelligence in this light}
\label{evaluatingIntelligenceInThisLight}

Earlier in this document, we have detailed how measuring skill alone does not move us forward when it comes to the development of broad abilities, we have suggested that AI evaluation should learn from its more mature sister field psychometrics (echoing the thesis of Psychometric AI and Universal Psychometrics), and we have provided a new formalism with practical implications for AI evaluation, pointing out the importance of the concept of scope, potential, generalization difficulty, experience, and priors. The following section summarizes key practical conclusions with respect to AI evaluation.

\subsubsection{Fair comparisons between intelligent systems}
\label{fairComparisons}

We mentioned in \ref{practicalImplications} that our formalism suggests concrete guidelines for comparing the intelligence of systems of different nature, such as human intelligence and artificial intelligence. Being able to make such comparisons in a fair and rigorous way is essential to progress towards human-like general AI. Here we argue how such intelligence comparisons between systems entail specific requirements with regard to the target systems' \textit{scope, potential}, and \textit{priors}. We also detail how such comparisons should proceed given that these requirements are met.

\paragraph*{Scope and potential requirements.}

In \ref{theMeaningOfGenerality}, we argued that intelligence is necessarily tied to a scope of application, an idea also central to the formalism introduced in \ref{aFormalSynthesis}. As such, \textit{a comparison scale must be tied to a well-defined scope of tasks that is shared by the target systems} (all target systems should be able to learn to perform the same tasks).

Further, we must consider that the target systems may have different \textit{potential} (maximum achievable skill) over their shared scope. An intelligence comparison should focus on skill-acquisition efficiency, but skill-acquisition efficiency cannot be meaningfully compared between systems that arrive at vastly different levels of skills. As such, \textit{a comparison scale must be tied to a fixed threshold of skill over the scope of tasks considered}. This skill threshold should be achievable by all target systems.

For instance, comparing a generally-intelligent system to human intelligence would only make sense if the scope of tasks that can be learned by the system is the same scope of tasks that can be learned by a typical human, and the comparison should focus on the efficiency with which the system achieves the same level of skill as a human expert. Comparing maximum realized skill does not constitute an intelligence comparison.

\paragraph*{Prior knowledge requirements.}

Since the formalism of \ref{aFormalSynthesis} summarizes priors into a single scalar score, which is homogeneous to the score used to quantify experience, it is not strictly necessary for the two systems being compared to share the same priors. For instance, if two systems achieve the same skill using the same amount of experience (the exact nature of this experience, determined by the curriculum used, may differ), the system that has the least amount of prior knowledge would be considered more intelligent.

However, it would be generally impractical to fully quantify prior knowledge. As such, \textit{we recommend only comparing the intelligence of systems that assume a sufficiently similar set of priors}. This implies that any measure of intelligence should explicitly and exhaustively list the priors it assumes, an idea we detail below, in \ref{whatToExpectOfAnIdealBenchmark}. Further, this implies that systems that aim at implementing human-like general intelligence should leverage Core Knowledge priors.\\

If the above conditions are met (shared scope, well-defined skill threshold over scope, and comparable knowledge priors), then a fair intelligence comparison would then consist of contrasting the skill-acquisition efficiency profile of the target systems. The more intelligent system would be the one that uses the least amount of experience to arrive at the desired skill threshold in the average case. Alternatively, computation efficiency, energy efficiency, and risk efficiency may also be considered, as per \ref{otherEfficiencies}.

\subsubsection{What to expect of an ideal intelligence benchmark}
\label{whatToExpectOfAnIdealBenchmark}

The recommendations below synthesizes the conclusions of this document with regard to the properties that a candidate benchmark of human-like general intelligence should possess.

\begin{itemize}
    \item It should describe its scope of application and its own predictiveness with regard to this scope (i.e. it should establish \textit{validity}). In practice, this would be achieved by empirically determining the statistical relationship between success on the benchmark and success on a range of real-world tasks.
    \item It should be reliable (i.e. \textit{reproducible}). If an evaluation session includes stochastic elements, sampling different values for these elements should not meaningfully affect the results. Different researchers independently evaluating the same system or approach using the benchmark should arrive at the same conclusions.
    \item It should set out to measure broad abilities and developer-aware generalization:
        \begin{itemize}
            \item It should not be solely measuring skill or potential (maximum achievable skill).
            \item It should not feature in its evaluation set any tasks that are known in advance, either to the test-taking system itself or to the developers of the system (cf. developer-aware generalization as defined in \ref{spectrumOfGeneralization}).
            \item It should seek to quantify the generalization difficulty it measures (cf. formal definition from \ref{aFormalSynthesis}), or at least provide qualitative guidelines with regard to its generalization difficulty: it should at least be made clear whether the benchmark seeks to measure local generalization (robustness), broad generalization (flexibility), or extreme generalization (general intelligence), as defined in \ref{spectrumOfGeneralization}. Taking into account generalization difficulty minimizes the possibility that a given benchmark could be ``hacked'' by solvers that take undesired shortcuts that bypass broad abilities (e.g. leveraging surface textures instead of semantic content in image recognition).
        \end{itemize}
    \item It should control for the amount of experience leveraged by test-taking systems during training. It should not be possible to ``buy'' performance on the benchmark by sampling unlimited training data. The benchmark should avoid tasks for which new data can be generated at will. It should be, in effect, a game for which it is not possible to practice in advance of the evaluation session.
    \item It should explicitly and exhaustively describe the set of priors it assumes. Any task is going to involve priors, but in many tasks used for AI evaluation today, priors stay implicit, and the existence of implicit hidden priors may often give an unfair advantage to either humans or machines.
    \item It should work for both humans and machines, fairly, by only assuming the same priors as possessed by humans (e.g. Core Knowledge) and only requiring a human-sized amount of practice time or training data.
\end{itemize}

These recommendations for general AI evaluation wouldn't be complete without a concrete effort to implement them. In part III, we present our initial attempt.


\section{A benchmark proposal: the ARC dataset}
\label{theARCDataset}

In this last part, we introduce the Abstraction and Reasoning Corpus (ARC), a dataset intended to serve as a benchmark for the kind of general intelligence defined in \ref{aFormalSynthesis}. ARC is designed to incorporate as many of the recommendations of \ref{evaluatingIntelligenceInThisLight} as possible.

\subsection{Description and goals}
\label{descritionAndGoals}

\subsubsection{What is ARC?}
\label{whatIsARC}

ARC can be seen as a general artificial intelligence benchmark, as a program synthesis benchmark, or as a psychometric intelligence test. It is targeted at both humans and artificially intelligent systems that aim at emulating a human-like form of general fluid intelligence. It is somewhat similar in format to Raven's Progressive Matrices \cite{Raven2003}, a classic IQ test format going back to the 1930s.\\
\\
ARC has the following top-level goals:

\begin{itemize}
    \item Stay close in format to psychometric intelligence tests (while addressing issues found in previous uses of such tests for AI evaluation, as detailed in \ref{ARCKeyDifferencesWithPsychometricTests}), so as to be approachable by both humans and machines; in particular it should be solvable by humans without any specific practice or training.
    \item Focus on measuring developer-aware generalization, rather than task-specific skill, by only featuring novel tasks in the evaluation set (assumed unknown to the developer of a test-taker).
    \item Focus on measuring a qualitatively ``broad'' form of generalization (cf. \ref{spectrumOfGeneralization}), by featuring highly abstract tasks that must be understood by a test-taker using very few examples.
    \item Quantitatively control for experience by only providing a fixed amount of training data for each task and only featuring tasks that do not lend themselves well to artificially generating new data.
    \item Explicitly describe the complete set of priors it assumes (listed in \ref{ARCCoreKnowledge}), and enable a fair general intelligence comparison between humans and machines by only requiring priors close to innate human prior knowledge (cf. \ref{whatToExpectOfAnIdealBenchmark}).

\end{itemize}

ARC comprises a training set and an evaluation set. The training set features 400 tasks, while the evaluation set features 600 tasks. The evaluation set is further split into a public evaluation set (400 tasks) and a private evaluation set (200 tasks). All tasks are unique, and the set of test tasks and the set of training tasks are disjoint. The task data is available at \href{https://github.com/fchollet/ARC}{github.com/fchollet/ARC}.

\begin{figure}[h]
    \centering
    \includegraphics[scale=0.46]{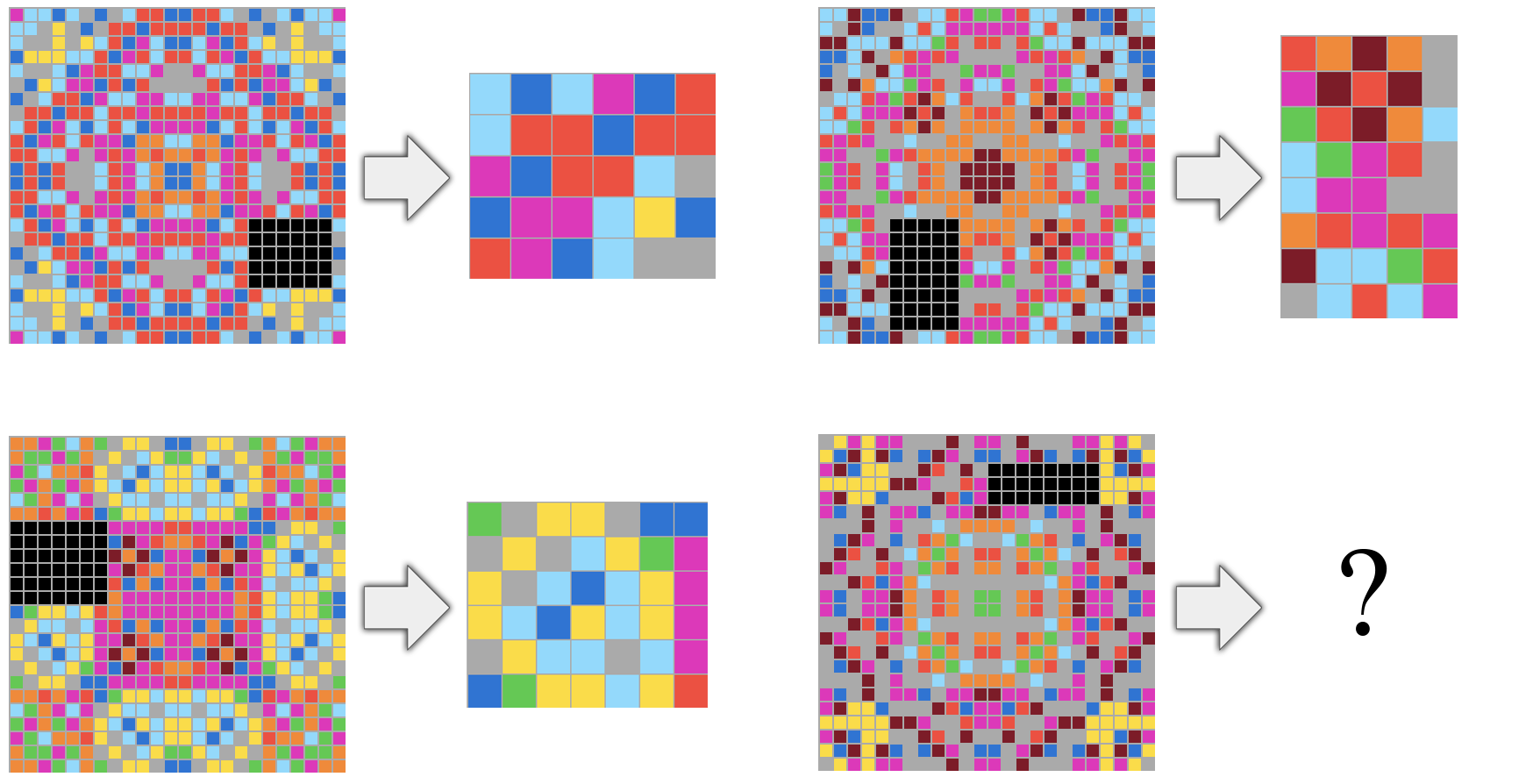}
    \caption{A task where the implicit goal is to complete a symmetrical pattern. The nature of the task is specified by three input/output examples. The test-taker must generate the output grid corresponding to the input grid of the test input (bottom right).}
    \label{fig:arc-example}
\end{figure}

Each task consists of a small number of demonstration examples (3.3 on average), and a small number of test examples (generally 1, although it may be 2 or 3 in rare cases). Each example consists of an ``input grid'' and an ``output grid''. Each ``grid'' is a literal grid of symbols (each symbol is typically visualized via a unique color), as seen in figure \ref{fig:arc-example}. There are 10 unique symbols (or colors). A grid can be any height or width between 1x1 and 30x30, inclusive (the median height is 9 and the median width is 10).

When solving an evaluation task, a test-taker has access to the training examples for the task (both the input and output grids), as well as the input grid of the test examples for the task. The test-taker must construct on its own the output grid corresponding to the input grid of each test example. ``Constructing the output grid'' is done entirely from scratch, meaning that the test-taker must decide what the height and width of the output grid should be, what symbols it should place on the grid, and where. The task is successfully solved if the test-taker can produce the exact correct answer on all test examples for the task (binary measure of success). For each test example in a task, the test-taker (either human or machine) is allowed 3 trials \footnote{We consider 3 trials to be enough to account for cases in which the task may be slightly ambiguous or in which the test-taker may commit mechanical errors when inputting an answer grid.}. The only feedback received after a trial is binary (correct answer or incorrect answer). The score of an intelligent system on ARC is the fraction of tasks in the evaluation set that it can successfully solve. Crucially, it is assumed that neither the test-taker nor its developer would have had any prior information about the tasks featured in the evaluation set: ARC seeks to measure ``developer aware generalization'' as defined in \ref{spectrumOfGeneralization}. The existence of a private evaluation set enables us to strictly enforce this in the setting of a public competition.

A test-taker is also assumed to have access to the entirety of the training set, although the training data isn't strictly necessary in order to be successful on the validation set, as all tasks are unique and do not assume any knowledge other than the priors described in \ref{ARCCoreKnowledge}. A typical human can solve most of the ARC evaluation set without any previous training. As such, the purpose of the training set primarily to serve as a development validation set for AI system developers, or as a mock test for human test-takers. It could also be used as a way to familiarize an algorithm with the content of Core Knowledge priors. We do not expect that practice on the training set would increase human performance on the test set (albeit this hypothesis would need to be concretely tested).

\subsubsection{Core Knowledge priors}
\label{ARCCoreKnowledge}

Any test of intelligence is going to involve prior knowledge. ARC seeks to control for its own assumptions by explicitly listing the priors it assumes, and by avoiding reliance on any information that isn't part of these priors (e.g. acquired knowledge such as language). The ARC priors are designed to be as close as possible to Core Knowledge priors, so as to provide a fair ground for comparing human intelligence and artificial intelligence, as per our recommendations in \ref{fairComparisons}.
\\
\\
The Core Knowledge priors assumed by ARC are as follows:

\subsubsection*{a. Objectness priors:}

\begin{figure}[h]
    \centering
    \includegraphics[scale=0.3]{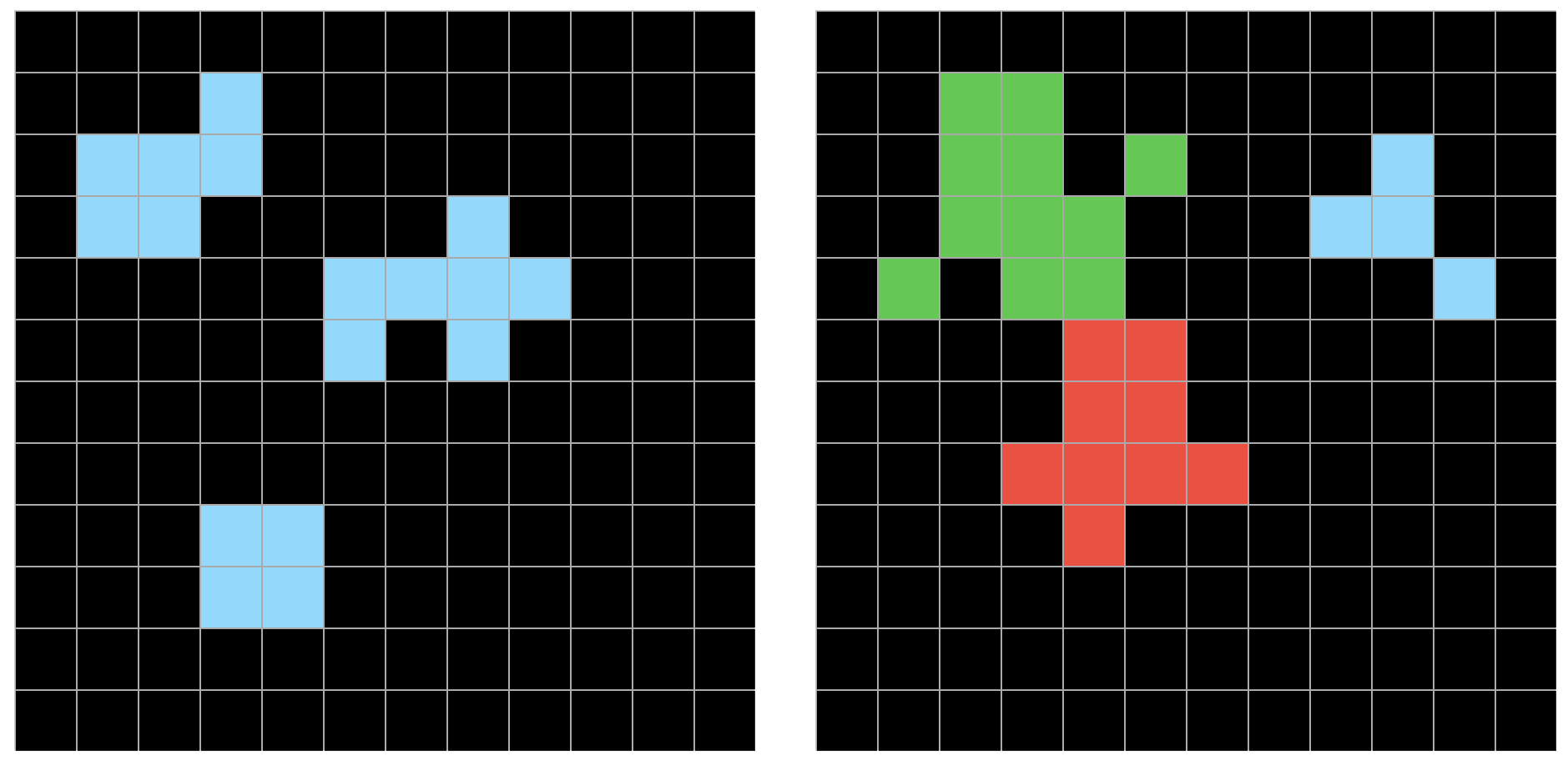}
    \caption{Left, objects defined by spatial contiguity. Right, objects defined by color continuity.}
    \label{fig:objectness-by-color-or-contiguity}
\end{figure}

\begin{figure}[h]
    \centering
    \includegraphics[scale=0.5]{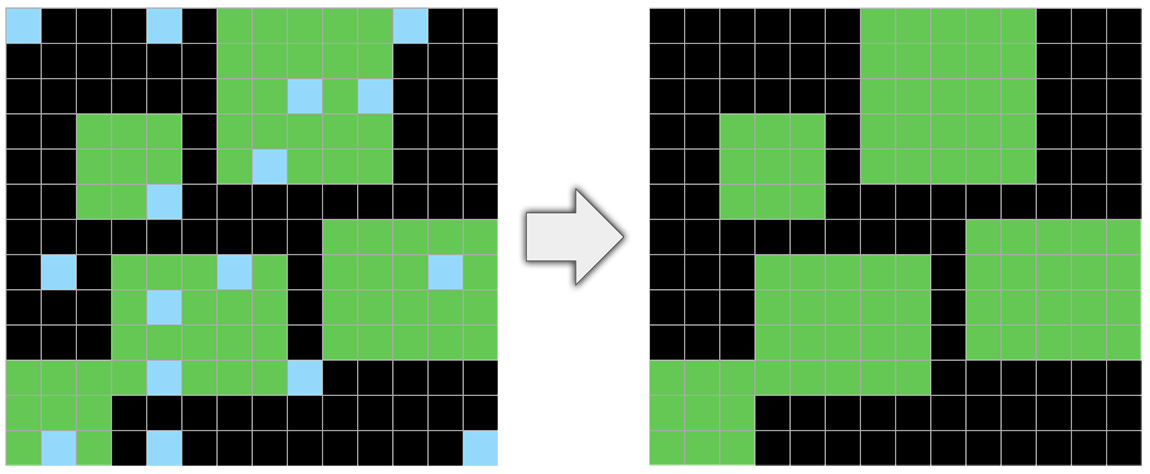}
    \caption{A denoising task.}
    \label{fig:denoising-task}
\end{figure}

\begin{figure}[h]
    \centering
    \includegraphics[scale=0.41]{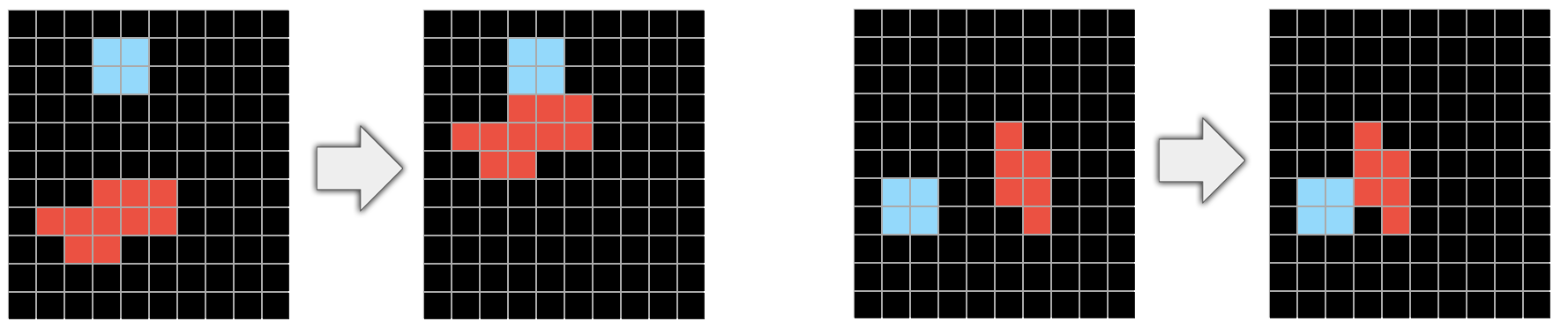}
    \caption{The red object ``moves'' towards the blue object until ``contact''.}
    \label{fig:translate-until-contact}
\end{figure}

\begin{figure}[h]
    \centering
    \includegraphics[scale=0.40]{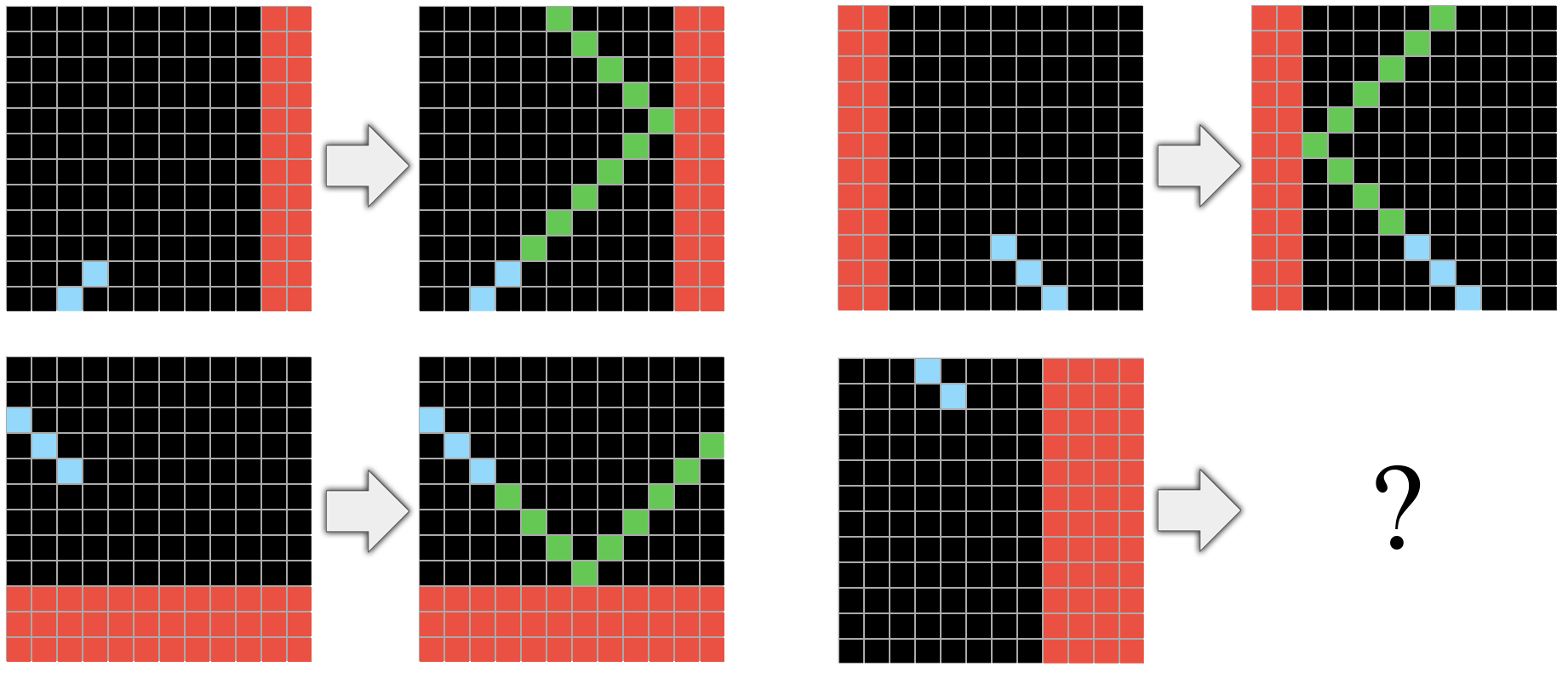}
    \caption{A task where the implicit goal is to extrapolate a diagonal line that ``rebounds'' upon contact with a red obstacle.}
    \label{fig:line-rebounding}
\end{figure}

\textbf{Object cohesion:} Ability to parse grids into ``objects'' based on continuity criteria including color continuity or spatial contiguity (figure \ref{fig:objectness-by-color-or-contiguity}), ability to parse grids into zones, partitions.
\\\\
\textbf{Object persistence:} Objects are assumed to persist despite the presence of noise (figure \ref{fig:denoising-task}) or occlusion by other objects. In many cases (but not all) objects from the input persist on the output grid, often in a transformed form. Common geometric transformations of objects are covered in category 4, ``basic geometry and topology priors''.
\\\\
\textbf{Object influence via contact:} Many tasks feature physical contact between objects (e.g. one object being translated until it is in contact with another (figure \ref{fig:translate-until-contact}), or a line ``growing'' until it ``rebounds'' against another object (figure \ref{fig:line-rebounding}).

\subsubsection*{b. Goal-directedness prior:}

\begin{figure}[h]
    \centering
    \includegraphics[scale=0.40]{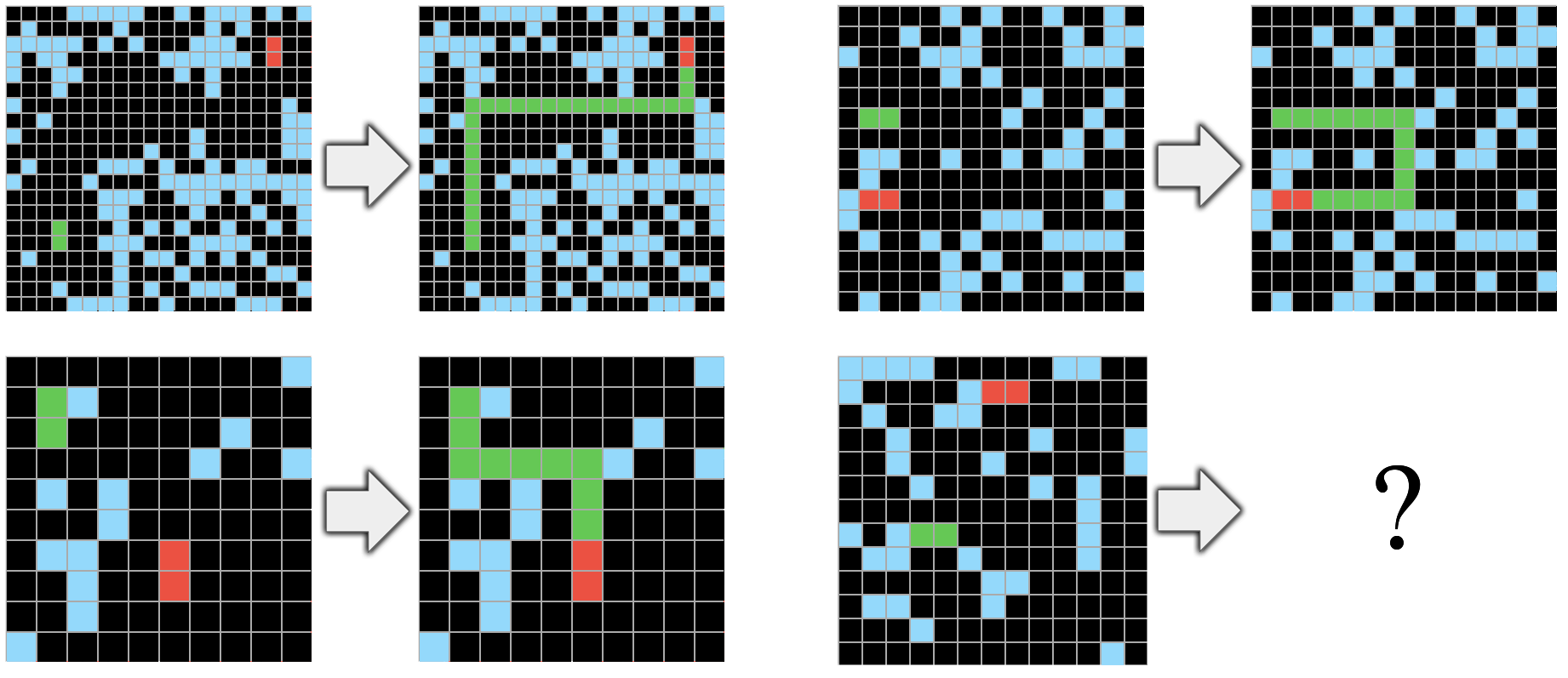}
    \caption{A task that combines the concepts of ``line extrapolation'', ``turning on obstacle'', and ``efficiently reaching a goal'' (the actual task has more demonstration pairs than these three).}
    \label{fig:efficiently-reaching-goal}
\end{figure}

While ARC does not feature the concept of time, many of the input/output grids can be effectively modeled by humans as being the starting and end states of a process that involves intentionality (e.g. figure \ref{fig:efficiently-reaching-goal}). As such, the goal-directedness prior may not be strictly necessary to solve ARC, but it is likely to be useful.

\subsubsection*{c. Numbers and Counting priors:}

\begin{figure}[h]
    \centering
    \includegraphics[scale=0.40]{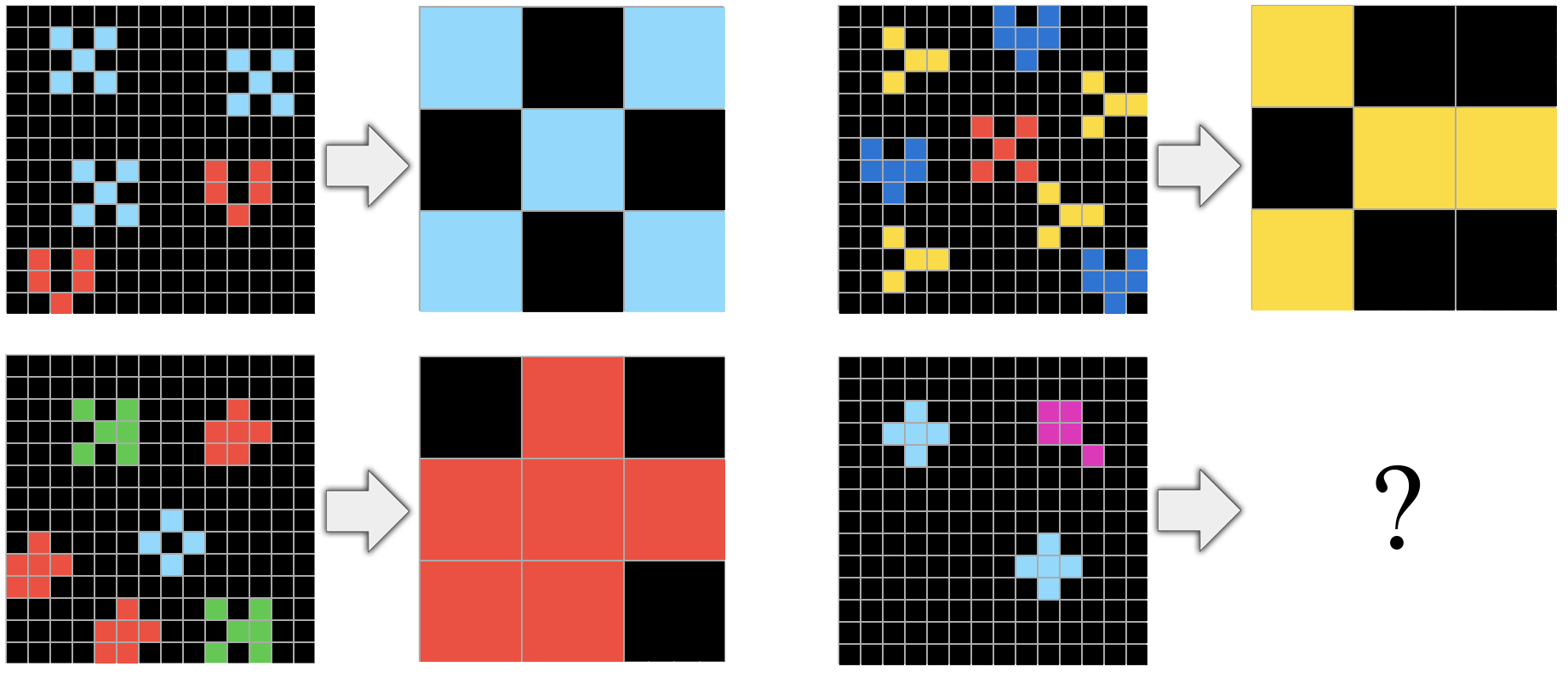}
    \caption{A task where the implicit goal is to count unique objects and select the object that appears the most times (the actual task has more demonstration pairs than these three).}
    \label{fig:counting_objects}
\end{figure}

Many ARC tasks involve counting or sorting objects (e.g. sorting by size), comparing numbers (e.g. which shape or symbol appears the most (e.g. figure \ref{fig:counting_objects})? The least? The same number of times? Which is the largest object? The smallest? Which objects are the same size?), or repeating a pattern for a fixed number of time. The notions of addition and subtraction are also featured (as they are part of the Core Knowledge number system as per \cite{Spelke2007}). All quantities featured in ARC are smaller than approximately 10.

\subsubsection*{d. Basic Geometry and Topology priors:}

\begin{figure}[h]
    \centering
    \includegraphics[scale=0.40]{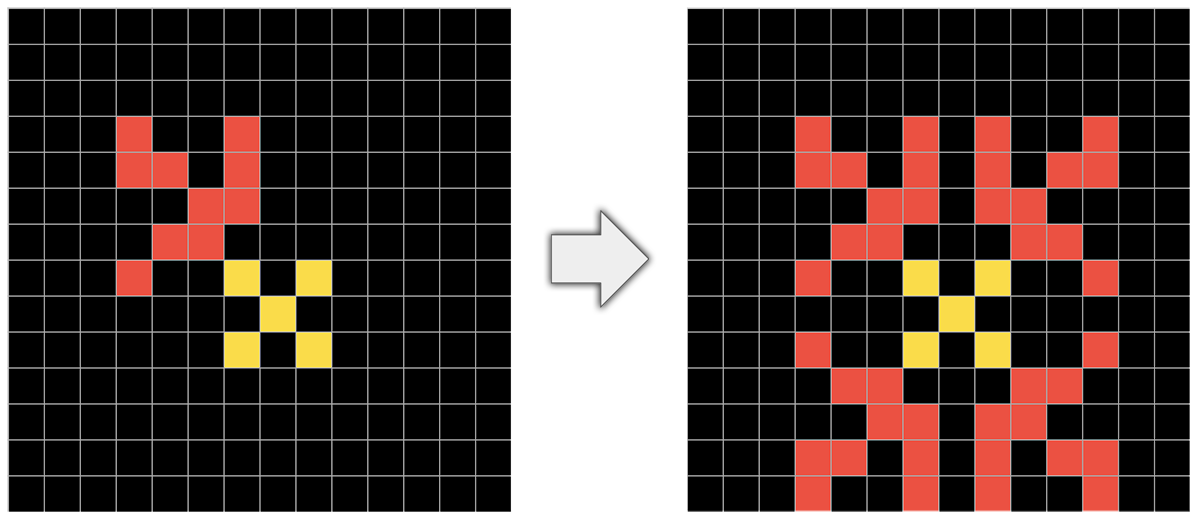}
    \caption{Drawing the symmetrized version of a shape around a marker. Many tasks involve some form of symmetry.}
    \label{fig:draw-symmetrized-version}
\end{figure}

ARC tasks feature a range of elementary geometry and topology concepts, in particular:

\begin{itemize}
    \item Lines, rectangular shapes (regular shapes are more likely to appear than complex shapes).
    \item Symmetries (e.g. figure \ref{fig:draw-symmetrized-version}), rotations, translations.
    \item Shape upscaling or downscaling, elastic distortions.
    \item Containing / being contained / being inside or outside of a perimeter.
    \item Drawing lines, connecting points, orthogonal projections.
    \item Copying, repeating objects.
\end{itemize}

\subsubsection{Key differences with psychometric intelligence tests}
\label{ARCKeyDifferencesWithPsychometricTests}

We have pointed out in \ref{integrationAIEvaluationAndPsychometric} the reasons why using existing psychometric intelligence tests (or ``IQ tests'') does not constitute a sound basis for AI evaluation. Albeit ARC stays deliberately close in format to traditional IQ tests (as well as related efforts such as  Hern{\'a}ndez-Orallo's C-Test \cite{HO98}), its design differs from them in fundamental ways. We argue that these differences address the shortcomings of psychometric intelligence tests in the context of AI evaluation. In particular:

\begin{itemize}
    \item Unlike some psychometric intelligence tests, ARC is not interested in assessing crystallized intelligence or crystallized cognitive abilities. ARC only assesses a general form of fluid intelligence, with a focus on reasoning and abstraction. ARC does not involve language, pictures of real-world objects, or real-world common sense. ARC seeks to only involve knowledge that stays close to Core Knowledge priors, and avoids knowledge that would have to be acquired by humans via task-specific practice.
    \item The tasks featured in the ARC evaluation set are unique and meant to be unknown to developers of test-taking systems (as ARC seeks to assess developer-aware generalization). This prevents developers from solving the tasks themselves and hard-coding their solution in program form. This can be strictly enforced in competition settings via the existence of a private evaluation set.
    \item ARC has greater task diversity than typical psychometric intelligence tests (hundreds of unique tasks with limited overlap between tasks), which reduces the likelihood that hard-coding task-specific solutions would represent a practical shortcut for developers, even for the public evaluation set.
    \item Unlike tasks from the C-Test \cite{HO98}, ARC tasks are in majority not programmatically generated. We perceive programmatic generation from a static ``master'' program as a weakness, as it implies that merely reverse-engineering the generative program shared across tasks (presumably a simple program, since it had to be written down by the test developer) would be sufficient to fully solve all tasks. Manual task generation increases task diversity and reduces the risk of existence of an unforeseen shortcut that could be used to by-pass the need for broad abilities in solving the test.

\end{itemize}

\subsubsection{What a solution to ARC may look like, and what it would imply for AI applications}
\label{ARCWhatASolutionMayLookLike}

We have found ARC to be fully solvable by humans. While many ARC test tasks are intellectually challenging, human test-takers appear to be able to solve the majority of tasks on their first try without any practice or verbal explanations. Each task included in ARC has been successfully solved by at least one member of a group of three high-IQ humans (who did not communicate with each other), which demonstrates task feasibility. In the future, we hope to be able to further investigate human performance on ARC by gathering a statistically significant amount of human testing data, in particular with regard to the relationship between CHC cognitive abilities and ARC performance.

Crucially, to the best of our knowledge, ARC does not appear to be approachable by any existing machine learning technique (including Deep Learning), due to its focus on broad generalization and few-shot learning, as well as the fact that the evaluation set only features tasks that do not appear in the training set.

For a researcher setting out to solve it, ARC is perhaps best understood as a program synthesis benchmark. Program synthesis \cite{gulwani2015inductive, gulwani2017program} is a subfield of AI interested in the generation of programs that satisfy a high-level specification, often provided in the form of pairs of example inputs and outputs for the program -- which is exactly the ARC format.

A hypothetical ARC solver may take the form of a program synthesis engine that uses the demonstration examples of a task to generate candidates that transform input grids into corresponding output grids. Schematically:

\begin{itemize}
    \item Start by developing a domain-specific language (DSL) capable of expressing all possible solution programs for any ARC task. Since the exact set of ARC tasks is purposely not formally definable, this may be challenging (the space of tasks is defined as \textit{anything} expressible in terms of ARC pairs that would only involve Core Knowledge). It would require harding-coding the Core Knowledge priors from \ref{ARCCoreKnowledge} in a sufficiently abstract and combinable program form, to serve as basis functions for a kind of ``human-like reasoning DSL''. We believe that solving this specific subproblem is critical to general AI progress.
    \item Given a task, use the DSL to generate a set of candidate programs that turn the inputs grids into the corresponding output grids. This step would reuse and recombine subprograms that previously proved useful in other ARC tasks.
    \item Select top candidates among these programs based on a criterion such as program simplicity or program likelihood (such a criterion may be trained on solution programs previously generated using the ARC training set). Note that we do not expect that merely selecting the simplest possible program that works on training pairs will generalize well to test pairs (cf. our definition of generalization difficulty from \ref{aFormalSynthesis}).
    \item Use the top three candidates to generate output grids for the test examples.

\end{itemize}

We posit that the existence of a human-level ARC solver would represent the ability to program an AI from demonstrations alone (only requiring a handful of demonstrations to specify a complex task) to do a wide range of human-relatable tasks of a kind that would normally require human-level, human-like fluid intelligence. As supporting evidence, we note that human performance on psychometric intelligence tests (which are similar to ARC) is predictive of success across all human cognitive tasks. Further, we posit that, since an ARC solver and human intelligence would both be founded on the same knowledge priors, the scope of application of an ARC solver would be close to that of human cognition, making such a solver both practically valuable (i.e. it could solve useful, human-relevant problems) and easy to interact with (i.e. it would readily understand human demonstrations and would produce behavior that is in line with human expectations).

Our claims are highly speculative and may well prove fully incorrect, much like Newell's 1973 hopes that progress on chess playing would translate into meaningful progress on achieving a range of broad cognitive abilities -- especially if ARC turns out to feature unforeseen vulnerabilities to unintelligent shortcuts. We expect our claims to be validated or invalidated in the near future once we make sufficient progress on solving ARC.

\subsection{Weaknesses and future refinements}
\label{ARCWeaknessesAndFutureRefinements}

It is important to note that ARC is a work in progress, not a definitive solution; it does not fit all of the requirements listed in \ref{whatToExpectOfAnIdealBenchmark}, and it features a number of key weaknesses:

\begin{itemize}
    \item \textbf{Generalization is not quantified.} While ARC is explicitly designed to measure ``broad generalization'' as opposed to ``local generalization'' or ``extreme generalization'', we do not offer a quantitative measure of the generalization of the evaluation set given the test set, or the generalization difficulty of each task (considered independently). We plan on conducting future work to empirically address this issue by using human performance on a task (considered over many human subjects) to estimate the generalization difficulty it represents. We would be particularly interested in attempting to correlate human performance on a task with an approximation of the AIT measure of generalization difficulty proposed in \ref{aFormalSynthesis} (such an approximation should become available as we make progress on ARC solver programs). Finding high correlation, or a lack of correlation, would provide a degree of validation or invalidation of our formal measure.
    \item \textbf{Test validity is not established.} Validity represents the predictiveness of test performance with regard to performance on other non-test activities. The validity of ARC should be investigated via large-sample size statistical studies on humans, following the process established by psychometrics. Further, when AI ARC solvers become a reality, we will also be able to study how well ARC performance translates into real-world usefulness across a range of tasks.
    \item \textbf{Dataset size and diversity may be limited.} ARC only features 1,000 tasks in total, and there may be some amount of conceptual overlap across many tasks. This could make ARC potentially vulnerable to shortcut strategies that could solve the tasks without featuring intelligence. We plan on running public AI competitions (using the private evaluation set) as a way to crowd-source attempts to produce such shortcuts (if a shortcut exists, it should arise quickly in a competition setting). Further, to mitigate potential vulnerability against such shortcuts, we intend to keep adding new tasks to ARC in the future, possibly by crowd-sourcing them.
    \item \textbf{The evaluation format is overly close-ended and binary.} The score of a test-taker on an evaluation task is either 0 or 1, which lacks granularity. Further, real-world problem-solving often takes the form of an interactive process where hypotheses are formulated by the test-taker then empirically tested, iteratively. In ARC, this approach is possible to an extent since the test-taker is allowed 3 trials for each test example in a task. However, this format remains overly limiting. A better approach may be let the test taker dynamically interact with an example generator for the task: the test taker would be able to ask for a new test input at will, would propose a solution for the test input, and would receive feedback on their solution, repeatedly, until the test-taker is reliably able to produce the correct answer. The test-taker's score on the task would then be a measure of the amount of feedback it required until it became able to reliably generate the correct solution for any new input. This represents a more direct measure of intelligence as formally defined in \ref{aFormalSynthesis}, where the input generator is in control of the curriculum.
    \item \textbf{Core Knowledge priors may not be well understood and may not be well captured in ARC.} Central to ARC is the notion that it only relies on innate human prior knowledge and does not feature significant amounts of acquired knowledge. However, the exact nature of innate human prior knowledge is still an open problem, and whether these priors are correctly captured in ARC is unclear.
\end{itemize}

\subsection{Possible alternatives}
\label{ARCPossibleAlternatives}

ARC is merely one attempt to create a human-like general intelligence benchmark that embodies as many of the guidelines listed in \ref{evaluatingIntelligenceInThisLight} as possible. While ARC stays very close to the format of psychometric intelligence tests, many other possible approaches could be explored. In this section, we offer some suggestions for alternatives.

\subsubsection{Repurposing skill benchmarks to measure broad generalization}

We noted in \ref{currentTrendsInAIEvaluation} the ongoing fascination of the AI research community in developing systems that surpass human skill at board games and video games. We propose repurposing such tests of skills into tests of intelligence.

Consider an AI developer interested in solving game $X$. While the AI would be trained on instances of $X$, an evaluation arbiter would create multiple variants of $X$ ($X_1$, $X_2$, … $X_n$). These alternative games would be designed to represent a meaningful amount of generalization difficulty over $X$ (as defined in \ref{aFormalSynthesis}): the simplest game-playing program that is optimal on instances of $X$ (e.g. game levels of $X$) would not be optimal on $X_i$. As such, these alternative games would not be mere ``new levels'' of $X$, but would feature related-yet-novel gameplay, so as to measure broad generalization as opposed to local generalization. These alternative games would stay unknown to the AI developers, so as to measure developer-aware generalization. This proposed setup is thus markedly different from e.g. CoinRun \cite{CoinRun} or Obstacle Tower \cite{Juliani2019}, where the evaluation environments are not alternative games, but only levels of the same game (local generalization, or generalization to known unknowns), randomly sampled from a level generator which is known in advance to the AI developers (no evaluation of developer-aware generalization).

The AI trained on $X$, once ready, would then be tasked with learning to solve $X_1$, … $X_n$. Its evaluation score would then be a measure of the amount of experience it required on each alternative game in order to reach a specific threshold of skill, modulated by the amount of generalization difficulty represented by each alternative game. A measure of the general intelligence of such an AI would then be an average of these evaluation scores over a large number of different source games $X$.

For instance, consider the game \href{https://dota2.com}{DotA2}: an AI trained on DotA2 may be evaluated by measuring the efficiency with which it can learn to play new games from the same genre, such as \href{https://leagueoflegends.com}{League of Legends} or \href{https://heroesofthestorm.com}{Heroes of the Storm}. As an even simpler (but weaker) alternative, an AI trained on 16 specific DotA2 characters may be evaluated by measuring the efficiency with which it can learn to master a set of brand new characters it would not have played before -- for example, a strong human DotA2 player can play at a high level with a new character upon first try.

\subsubsection{Open-ended adversarial or collaborative approaches}

We have pointed out in \ref{ARCWeaknessesAndFutureRefinements} some of the key limitations of having to craft evaluation tasks manually: it is a labor-intensive process that makes it difficult to formally control for generalization difficulty, that could potentially result in a low-diversity set of tasks, and that is not easily scalable (although crowd-sourcing tasks may partially address this problem). The diversity and scalablility points are especially critical given that we need a constant supply of substantially new tasks in order to guarantee that the benchmark is measuring developer-aware generalization.

\begin{figure}[h]
    \centering
    \includegraphics[scale=0.26]{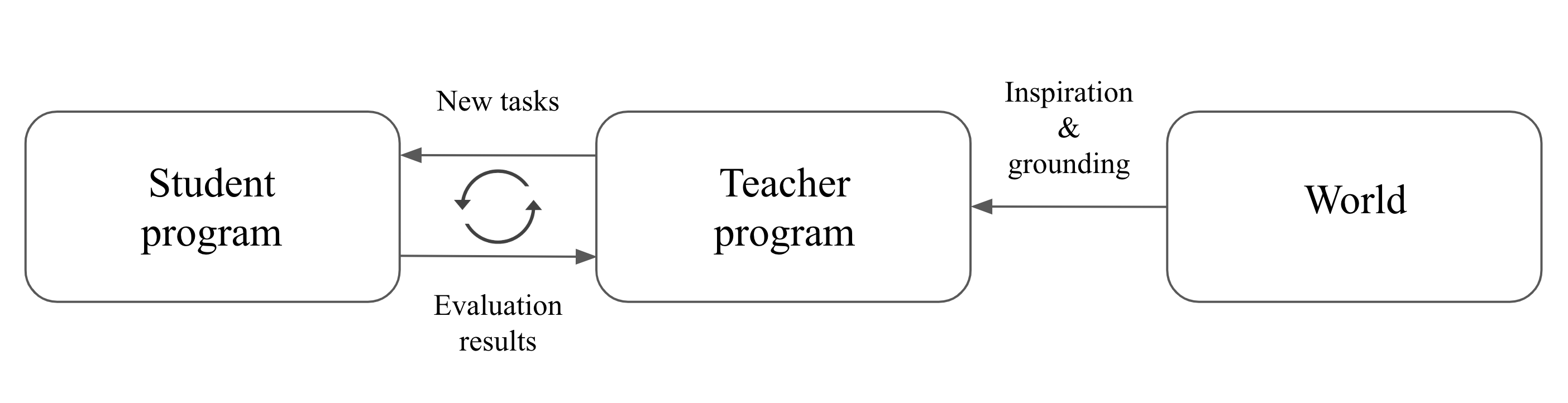}
    \caption{Teacher-student learning and evaluation system.}
    \label{fig:teacher-student-networks}
\end{figure}

A solution may be to instead programmatically generate new tasks. We noted in \ref{ARCKeyDifferencesWithPsychometricTests} that programmatic generation from a static ``master'' program is not desirable, as it places a ceiling on the diversity and complexity of the set of tasks that can be generated, and it offers a potential avenue to ``cheat'' on the benchmark by reverse-engineering the master program. We propose instead to generate tasks via an ever-learning program called a ``teacher'' program, interacting in a loop with test-taking systems, called ``student'' programs (figure \ref{fig:teacher-student-networks}). The teacher program would optimize task generation for novelty and interestingness for a given student (tasks should be new and challenging, while still being solvable by the student), while students would evolve to learn to solve increasingly difficult tasks. This setup is also favorable to curriculum optimization, as the teacher program may be configured to seek to optimize the learning efficiency of its students. This idea is similar to the ``anytime intelligence test'' proposed in \cite{hernandez2010measuring} and to the POET system proposed in \cite{Wang2019PairedOT}.

In order to make sure that the space of generated tasks retains sufficient complexity and novelty over time, the teacher program should draw information from an external source (assumed to feature incompressible complexity), such as the real world. This external source of complexity makes the setup truly open-ended. A teacher program that generates novel tasks that partially emulate human-relevant tasks would have the added advantage that it would guide the resulting student programs towards a form of intelligence that could transfer to real-world human-relevant problems.

\section*{Taking stock}
\label{takingStock}

The study of general artificial intelligence is a field still in its infancy, and we do not wish to convey the impression that we have provided a definitive solution to the problem of characterizing and measuring the intelligence held by an AI system. Rather, we have introduced a new perspective on defining and evaluating intelligence, structured around the following ideas:

\begin{itemize}
    \item Intelligence is the efficiency with which a learning system turns experience and priors into skill at previously unknown tasks.
    \item As such, a measure of intelligence must account for priors, experience, and generalization difficulty.
    \item All intelligence is relative to a scope of application. Two intelligent systems may only be meaningfully compared within a shared scope and if they share similar priors.
    \item As such, general AI should be benchmarked against human intelligence and should be founded on a similar set of knowledge priors (e.g. Core Knowledge).

\end{itemize}

We also have provided a new formalism based on Algorithmic Information Theory (cf. \ref{aFormalSynthesis}) to rigorously and quantitatively reason about these ideas, as well as a set of concrete guidelines to follow for developing a benchmark of general intelligence (cf. \ref{fairComparisons} and \ref{whatToExpectOfAnIdealBenchmark}).

Our definition, formal framework, and evaluation guidelines, which do \textit{not} capture all facets of intelligence, were developed to be actionable, explanatory, and quantifiable, rather than being descriptive, exhaustive, or consensual. They are not meant to invalidate other perspectives on intelligence, rather, they are meant to serve as a useful objective function to guide research on broad AI and general AI, as outlined in \ref{practicalImplications}. Our hope is for some part of the AI community interested in general AI to break out of a longstanding and ongoing trend of seeking to achieve raw skill at challenging tasks, given unlimited experience and unlimited prior knowledge.

To ground our ideas and enable others to build upon them, we are also providing an actual benchmark, the Abstraction and Reasoning Corpus, or ARC:

\begin{itemize}
    \item ARC takes the position that intelligence testing should control for scope, priors, and experience: every test task should be novel (measuring the ability to understand a new task, rather than skill) and should assume an explicit set of priors shared by all test-takers.
    \item ARC explicitly assumes the same Core Knowledge priors innately possessed by humans.
    \item ARC can be fully solved by humans, but cannot be meaningfully approached by current machine learning techniques, including Deep Learning.
    \item ARC may offer an interesting playground for AI researchers who are interested in developing algorithms capable of human-like broad generalization. It could also offer a way to compare human intelligence and machine intelligence, as we assume the same priors.

\end{itemize}

Importantly, ARC is still a work in progress, with known weaknesses listed in \ref{ARCWeaknessesAndFutureRefinements}. We plan on further refining the dataset in the future, both as a playground for research and as a joint benchmark for machine intelligence and human intelligence.\\

The measure of the success of our message will be its ability to divert the attention of some part of the community interested in general AI, away from surpassing humans at tests of skill, towards investigating the development of human-like broad cognitive abilities, through the lens of program synthesis, Core Knowledge priors, curriculum optimization, information efficiency, and achieving extreme generalization through strong abstraction.

\bibliography{scibib}

\bibliographystyle{Science}

\clearpage

\end{document}